%% file: main.tex
\documentclass[10pt,twocolumn,letterpaper]{article}

\usepackage[pagenumbers]{cvpr} 

\input{preamble}

\definecolor{cvprblue}{rgb}{0.21,0.49,0.74}
\usepackage[pagebackref,breaklinks,colorlinks,citecolor=cvprblue]{hyperref}

\newcommand*{\affaddr}[1]{#1} 
\newcommand*{\affmark}[1][*]{\textsuperscript{#1}}
\newcommand*{\email}[1]{\texttt{#1}}

\usepackage{lipsum}
\def\paperID{*****} 
\def\confName{CVPR}
\def\confYear{2024}

\title{DynamiCrafter: Animating Open-domain Images with Video Diffusion Priors}

\author{%
Jinbo Xing\affmark[1]~~~ 
Menghan Xia\affmark[2,]$^*$~~~
Yong Zhang\affmark[2]~~~
Haoxin Chen\affmark[2]~~~
Wangbo Yu\affmark[3]~~~\\
Hanyuan Liu\affmark[1]~~~
Xintao Wang\affmark[2]~~~
Tien-Tsin Wong\affmark[1,]$^*$~~~
Ying Shan\affmark[2]\\
\affaddr{\affmark[1]The Chinese University of Hong Kong~~~~~~~~~~}
\affaddr{\affmark[2]Tencent AI Lab~~~~~~~~~~}
\affaddr{\affmark[3]Peking University~~~~~~~~~~}\\
\small\email{Project page: \url{https://doubiiu.github.io/projects/DynamiCrafter}}
}

\begin{document}
\maketitle
\addtocontents{toc}{\setcounter{tocdepth}{0}}
\input{tex/0_abstract} 
\footnotetext[1]{~Corresponding Authors.}
\input{tex/1_introduction}
\input{tex/2_relatedwork}
\input{tex/3_method}
\input{tex/4_experiment}
\input{tex/5_discussion}

\input{tex/6_application}
\input{tex/8_conclusion}
\clearpage
{
    \small
    \bibliographystyle{ieeenat_fullname}
    \bibliography{main}
}

\input{X_suppl}

\end{document}

%% file: preamble.tex
\usepackage[dvipsnames]{xcolor}

\usepackage{multirow}

\newcommand{\best}[1]{\textbf{#1}}

\definecolor{Gray}{gray}{0.95}
\usepackage{colortbl}
\newcolumntype{a}{>{\columncolor{Gray}}c}
\newcommand{\mc}[2]{\multicolumn{#1}{c}{#2}}
\usepackage{bm}

\usepackage{pifont}   
\newcommand{\cmark}{\ding{51}}%
\newcommand{\xmark}{\ding{55}}%
\usepackage{caption}
\usepackage{wrapfig}

\usepackage{etoc}

%% file: tex/0_abstract.tex
\begin{abstract}
Animating a still image offers an engaging visual experience. Traditional image animation techniques mainly focus on animating natural scenes with stochastic dynamics (e.g. clouds and fluid) or domain-specific motions (e.g. human hair or body motions), and thus limits their applicability to more general visual content. To overcome this limitation, we explore the synthesis of dynamic content for open-domain images, converting them into animated videos. The key idea is to utilize the motion prior of text-to-video diffusion models by incorporating the image into the generative process as guidance.
Given an image, we first project it into a text-aligned rich context representation space using a query transformer, which facilitates the video model to digest the image content in a compatible fashion. However, some visual details still struggle to be preserved in the resultant videos. To supplement with more precise image information, we further feed the full image to the diffusion model by concatenating it with the initial noises.
Experimental results show that our proposed method can produce visually convincing and more logical \& natural motions, as well as higher conformity to the input image. Comparative evaluation demonstrates the notable superiority of our approach over existing competitors.
\end{abstract}

%% file: tex/1_introduction.tex
\section{Introduction}
\label{sec:introduction}

Image animation has been a longstanding challenge in the fields of computer vision, with the goal of converting still images into video counterparts that display natural dynamics while preserving the original appearance of the images. Traditional heuristic approaches primarily concentrate on synthesizing stochastic and oscillating motions~\cite{lee2018stochastic,li2023generative} or customizing for specific object categories~\cite{holynski2021animating,karras2023dreampose}. However, the strong assumptions imposed on these methods limit their applicability in general scenarios, such as animating open-domain images.
Recently, text-to-video (T2V) generative models have achieved remarkable success in creating diverse and vivid videos from textual prompts. This inspires us to investigate the potential of leveraging such powerful video generation capabilities for image animation.

Our key idea is to govern the video generation process of T2V diffusion models by incorporating a conditional image. However, achieving the goal of image animation is still non-trivial, as it requires both visual context understanding (essential for creating dynamics) and detail preservation.
Recent studies on multi-modal controllable video diffusion models, such as VideoComposer~\cite{wang2023videocomposer} and I2VGen-XL~\cite{I2VGen-XL}, have made preliminary attempts to enable video generation with visual guidance from an image. Unfortunately, both are incompetent for image animation due to their less comprehensive image injection mechanisms, which results in either abrupt temporal changes or low visual conformity to the input image (see Figure~\ref{fig:qualitative}).
To address this challenge, we propose a dual-stream image injection paradigm, comprised of text-aligned context representation and visual detail guidance, which ensures that the video diffusion model synthesizes detail-preserved dynamic content in a complementary manner. We call this approach \textit{DynamiCrafter}.

Given an image, we first project it into the text-aligned rich context representation space through a specially designed context learning network. Specifically, it consists of a pre-trained CLIP image encoder to extract text-aligned image features and a learnable query transformer to further promote its adaptation to the diffusion models. The rich context features are used by the model via cross attention layers, which will then be combined with the text-conditioned features through gated fusion.
In some extend, the learned context representation trades visual details with text alignment which helps facilitate semantic understanding of image context so that reasonable and vivid dynamics could be synthesized. To supplement more precise visual details, we further feed the full image to the diffusion model by concatenating it with the initial noise. This dual-stream injection paradigm guarantees both plausible dynamic content and visual conformity to the input image.

Extensive experiments are conducted to evaluate our proposed method, which demonstrates notable superiority over existing competitors and even comparable performance with the latest commercial demos (like Gen-2~\cite{Gen-2} and PikaLabs~\cite{PikaLabs}). Furthermore, we offer discussion and analysis on some insightful designs for diffusion model based image animation, such as the roles of different visual injection streams, the utility of text prompts and their potential for dynamics control, which may inspire follow-ups to push forward this line of technique.
Besides of image animation, \textit{DynamiCrafter} can be easily adapted to support applications like storytelling video generation, looping video generation, and generative frame interpolation.
Our contributions are summarized as follows:
\begin{itemize}
\itemindent=5pt
    \item We introduce an innovative approach for animating open-domain images by leveraging video diffusion prior, significantly outperforming contemporary competitors.
    \item We conduct a comprehensive analysis on the conditional space of text-to-video diffusion models and propose a dual-stream image injection paradigm to achieve the challenging goal of image animation.
    \item We pioneer the study of text-based motion control for open-domain image animation and demonstrate the proof of concept through preliminary experiments.
\end{itemize}

%% file: tex/2_relatedwork.tex
\vspace{-1mm}
\section{Related Work}
\label{sec:relatedwork}
\vspace{-1mm}
\subsection{Image Animation}
\label{subsec:image_animation}
Generating animation from still images is a heavily studied research area. Early physical simulation-based approaches~\cite{jhou2015animating,chuang2005animating} focus on simulating the motion of specific objects, 
resulting in low generalizability due to the independent modeling of each object category. To produce more realistic motion, reference-based methods~\cite{prashnani2017phase,okabe2009animating,siarohin2019first,cheng2020time,karras2023dreampose,siarohin2019animating,siarohin2021motion,wang2021latent} transfer motion or appearance information from reference signals, such as videos, to the synthesis process. Although they demonstrate better temporal coherence, the need for additional guidances limits their practical application. Additionally, a stream of works based on GAN~\cite{shaham2019singan,hinz2021improved,karras2020analyzing} can generate frames by perturbing initial latents or performing random walk in the latent vector space. However, the generated motion is not plausible since the animated frames are just a visualization of the possible appearance space without temporal awareness. Recently, (learned) motion prior-based methods~\cite{endo2019animating,holynski2021animating,mallya2022implicit,ni2023conditional,weng2019photo,zhao2022thin,xiao2023automatic,hu2022make} animate still images through explicit or implicit image-based rendering with estimated motion field or geometry priors. Similarly, video prediction~\cite{zhang2020dtvnet,xiong2018learning,babaeizadeh2018stochastic,li2018flow,xue2016visual,voleti2022mcvd,hu2023dynamic,hoppe2022diffusion,franceschi2020stochastic} predicts future video frames starting from single images by learning spatio-temporal priors from video data.

Although existing approaches has achieved impressive performance, they primarily focus on animating motions in curated domains, particularly stochastic~\cite{endo2019animating,jhou2015animating,blattmann2021ipoke,xue2018visual,okabe2009animating,chuang2005animating,lee2018stochastic,dorkenwald2021stochastic} and oscillating~\cite{li2023generative} motion. Furthermore, the animated objects are limited to specific categories, \eg, fluid~\cite{holynski2021animating,okabe2009animating,holynski2021animating,mahapatra2022controllable}, natural scenes~\cite{xiong2018learning,li2023generative,cheng2020time,jhou2015animating,shaham2019singan}, human hair~\cite{xiao2023automatic}, portraits~\cite{wang2021latent,wang2020imaginator,geng2018warp}, and bodies~\cite{wang2021latent,weng2019photo,karras2023dreampose,siarohin2021motion,blattmann2021understanding,bertiche2023blowing}. In contrast, our work proposes a generic framework for animating open-domain images with a wide range of content and styles, which is extremely challenging due to the overwhelming complexity and vast diversity.

\vspace{-0.5mm}
\subsection{Video Diffusion Models}
\label{subsec:video_diffusion_models}
\vspace{-0.5mm}
Diffusion models (DMs)~\cite{sohl2015deep,ho2020denoising} have recently shown unprecedented generative power in text-to-image (T2I) generation~\cite{ramesh2022hierarchical,nichol2022glide,saharia2022photorealistic,rombach2022high,he2023scalecrafter,zhang2023real}. To replicate this success to video generation, the first video diffusion model (VDM)~\cite{ho2022video} is proposed to model low-resolution videos using a space-time factorized U-Net in pixel space. Imagen-Video~\cite{ho2022imagenvideo} presents effective cascaded DMs with $\mathbf{v}$-prediction for generating high-definition videos. To reduce training costs, subsequent studies~\cite{wang2023modelscope,he2022latent,blattmann2023align,zhou2022magicvideo,wang2023lavie} are engaged in transferring T2I to text-to-video (T2V)~\cite{singer2022make,luo2023videofusion,ge2023preserve,zhang2023show1}, and learning VDMs in latent or hybrid-pixel-latent space.

Although these models can generate high-quality videos, they only accept text prompts as the sole semantic guidance, which can be vague and may not accurately reflect users' intention. Similar to adding controls in T2I~\cite{zhang2023adding,mou2023t2i,ye2023ipadapter,shi2023instancebooth}, introducing control signals in T2V, such as structure~\cite{esser2023structure,xing2023make}, pose~\cite{ma2023follow,zhang2023controlvideo}, and Canny edge~\cite{khachatryan2023text2video}, has been increasingly receiving much attention.
However, visual conditions in VDMs~\cite{tang2023any,yin2023dragnuwa}, such as RGB images, remain under-explored. Most recently and concurrently, image condition is examined in Seer~\cite{gu2023seer}, VideoComposer~\cite{wang2023videocomposer},  and I2VGen-XL~\cite{I2VGen-XL} for (text-)image-to-video synthesis.
However, they either focus on the curated domain, \ie, indoor objects~\cite{gu2023seer}, or fail to generate temporally coherent frames and realistic motions~\cite{wang2023videocomposer} and preserve visual details of the input image~\cite{I2VGen-XL} due to insufficient context understanding and loss of information of the input image. 
Moreover, recent proprietary T2V models~\cite{singer2022make,molad2023dreamix,villegas2023phenaki,yu2023magvit} have been demonstrated to be extensible to image-to-video synthesis. However, their results rarely adhere to the input image and suffers from the unrealistic temporal variation issue. Our approach is built upon text-conditioned VDMs to leverage their rich dynamic prior for animating open-domain images, by incorporating tailored designs for better semantic understanding and conformity to the input image.

%% file: tex/3_method.tex
\vspace{-1.5mm}
\section{Method}
\label{sec:method}
\vspace{-1.5mm}
\begin{figure*}[!t]
    \centering
    \includegraphics[width=1\linewidth]{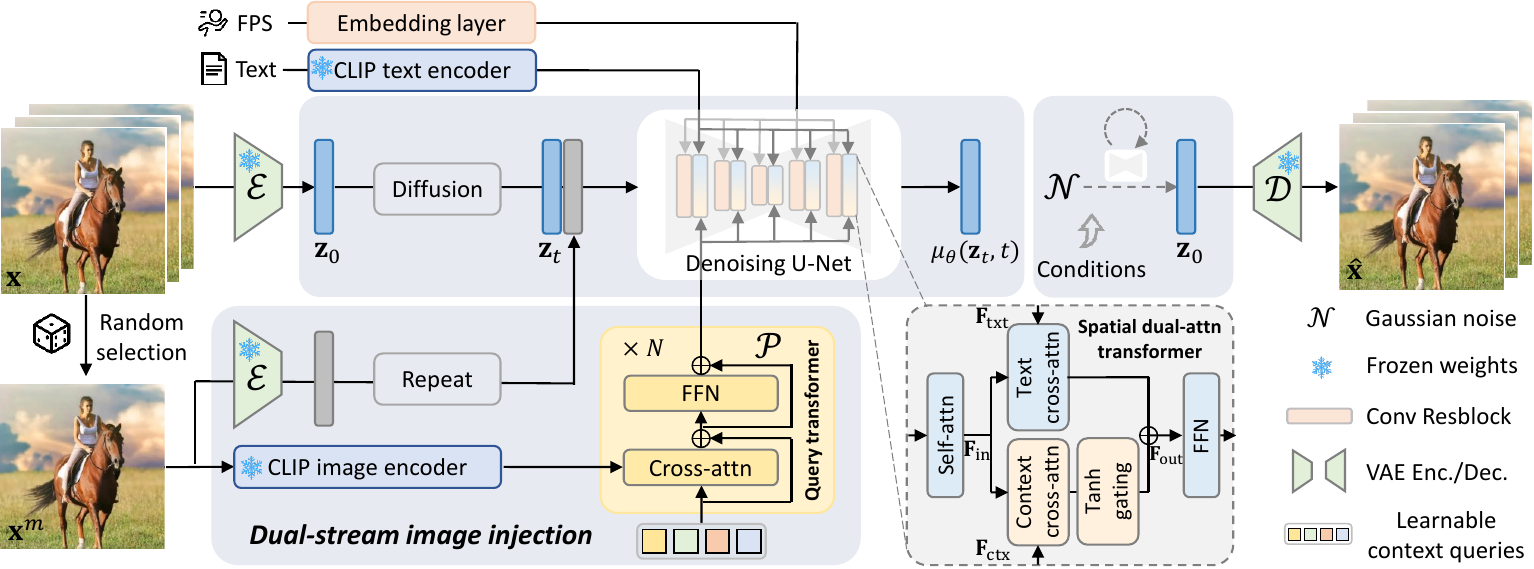}
    \vspace{-6mm}
    \caption{Flowchart of the proposed \emph{DynamiCrafter}. During training, we randomly select a video frame as the image condition of the denoising process through the proposed dual-stream image injection mechanism to inherit visual details and digest the input image in a context-aware manner. During inference, our model can generate animation clips from noise conditioned on the input still image.
    }
    \label{fig:overview}
    \vspace{-5.5mm}
\end{figure*}

Given a still image, we aim at animating it to produce a short video clip, that inherits all the visual content from the image and exhibits an implicitly suggested and natural dynamics.
Note that the still image can appear in the arbitrary location of the resultant frame sequence. 
Technically, such challenge can be formulated as a special kind of image-conditioned video generation that highly requires visual conformity. We tackle this synthesis task by utilizing the generative priors of pre-trained video diffusion models.

\subsection{Preliminary: Video Diffusion Models}
\label{subsec:preliminaries}

Diffusion models~\citep{ho2020denoising,sohldickstein2015deep} are generative models that define a forward diffusion process to convert data $\mathbf{x}_0 \sim p_{\text{data}}(\mathbf{x})$ into Gaussian noises $\mathbf{x}_T \sim \mathcal{N}(\mathbf{0}, \mathbf{I})$ and learn to reverse this process by denoising. The forward process $q(\mathbf{x}_t|\mathbf{x}_0,t)$ contains $T$ timesteps, which gradually adds noise to the data sample $\mathbf{x}_0$ to yield $\mathbf{x}_t$ through a parameterization trick. The denoising process $p_{\theta}(\mathbf{x}_{t-1}|\mathbf{x}_t,t)$ obtains less noisy data $\boldsymbol{x}_{t-1}$ from the noisy input $\boldsymbol{x}_t$ through a denoising network $\mathbf{\epsilon}_{\theta}\left(\mathbf{x}_t, t\right)$, which is supervised by the objective:
\begin{equation}
\min _{\theta} \mathbb{E}_{t, \mathbf{x} \sim p_{\text{data}}, \mathbf{\epsilon} \sim \mathcal{N}(\mathbf{0}, \mathbf{I})}\Vert\mathbf{\epsilon}-\mathbf{\epsilon}_{\theta}\left(\mathbf{x}_t, t\right)\Vert_2^2,
\end{equation} 
where $\mathbf{\epsilon}$ is the sampled ground truth noise and $\theta$ indicates the learnable network parameters. Once the model is trained, we can obtain denoised data $\mathbf{x}_0$ from a random noise $\mathbf{x}_T$ through iteratively denoising.

For video generation tasks, Latent Diffusion Models (LDMs)~\cite{ho2022imagenvideo} are commonly used to reduce the computation complexity. In this paper, our study is conducted based on an open-source video LDM \textit{VideoCrafter}~\cite{chen2023videocrafter1}. Given a video $\mathbf{x} \in \mathbb{R}^{L\times 3\times H\times W}$, we first encode it into a latent representation $\mathbf{z}=\mathcal{E}(\mathbf{x}), \mathbf{z} \in \mathbb{R}^{L\times C\times h\times w}$ frame-by-frame. Then, both the forward diffusion process $\mathbf{z}_t=p(\mathbf{z}_0, t)$ and backward denoising process $\mathbf{z}_t=p_{\theta}(\mathbf{z}_{t-1}, \mathbf{c}, t)$ are performed in this latent space, where $\mathbf{c}$ denotes possible denoising conditions like text prompt. Accordingly, the generated videos are obtained through the decoder $\hat{\mathbf{x}}=\mathcal{D}(\mathbf{z})$.

\subsection{Image Dynamics from Video Diffusion Priors}
\label{subsec:alignment}

An open-domain text-to-video diffusion model is assumed to have diverse dynamic visual content modeled conditioning on text descriptions.
To animate a still image with the T2V generative priors, the visual information should be injected into the video generation process in a comprehensive manner.
On the one hand, the image should be digested by the T2V model for context understanding, which is important for dynamics synthesis. On the other, the visual details should be preserved in the generated videos.
Based on this insight, we propose a dual-stream conditional image injection paradigm, consisting of text-aligned context representation and visual detail guidance. The overview diagram is illustrated in Figure~\ref{fig:overview}.

\vspace{-4mm}
\paragraph{Text-aligned context representation.}
To guide video generation with image context, we propose to project the image into a text-aligned embedding space, so that the video model can utilize the image information in a compatible fashion. 
Since the text embedding is constructed with pre-trained CLIP~\cite{radford2021learning} text encoder, we employ the image encoder counterpart to extract image feature from the input image. Although the global semantic token $\mathbf{f}_{\text{cls}}$ from the CLIP image encoder is well-aligned with image captions, it mainly represents the visual content at the semantic level and fails to capture the image's full extent.
To extract more complete information, we use the full visual tokens $\mathbf{F}_{\text{vis}}=\{\mathbf{f}^i\}^K_{i=1}$ from the last layer of the CLIP image ViT~\cite{dosovitskiy2020image}, which demonstrated high-fidelity in conditional image generation works~\cite{shi2023instancebooth,ye2023ipadapter}.
To promote the alignment with text embedding, in other words, to obtain a context representation that can be interpreted by the denoising U-Net, we utilize a learnable lightweight model $\mathcal{P}$ to translate $\mathbf{F}_{\text{vis}}$ into the final context representation $\mathbf{F}_{\text{ctx}}=\mathcal{P}(\mathbf{F}_{\text{vis}})$. We employ the query transformer architecture~\cite{jaegle2021perceiver,awadalla2023openflamingo} in multi-modal fusion studies as $\mathcal{P}$, which comprises $N$ stacked layers of cross-attention and feed-forward networks (FFN), and is adept at cross-modal representation learning via the cross-attention mechanism.

Subsequently, the text embedding $\mathbf{F}_{\text{txt}}$ and context embedding $\mathbf{F}_{\text{ctx}}$ are employed to interact with the U-Net intermediate features $\mathbf{F}_{\text{in}}$ through the dual cross-attention layers:
\begin{equation}
\mathbf{F}_{\text{out}} = \text{Softmax}(\frac{\mathbf{Q} \mathbf{K}_{\text{txt}}^{\top}}{\sqrt{d}})\mathbf{V}_{\text{txt}} + \lambda \cdot \text{Softmax}(\frac{\mathbf{Q} \mathbf{K}_{\text{ctx}}^{\top}}{\sqrt{d}})\mathbf{V}_{\text{ctx}},
\end{equation}
where $\mathbf{Q}=\mathbf{F}_{\text{in}}\mathbf{W}_\mathbf{Q}$, $\mathbf{K}_{\text{txt}}=\mathbf{F}_{\text{txt}}\mathbf{W}_\mathbf{K}$, $\mathbf{V}_{\text{txt}}=\mathbf{F}_{\text{txt}}\mathbf{W}_\mathbf{V}$, and $\mathbf{K}_{\text{ctx}}=\mathbf{F}_{\text{ctx}}\mathbf{W}'_\mathbf{K}$, $\mathbf{V}_{\text{ctx}}=\mathbf{F}_{\text{ctx}}\mathbf{W}'_\mathbf{V}$ accordingly.
In particular, $\lambda$ denotes the coefficient that fuses text-conditioned and image-conditioned features, which is achieved through \texttt{tanh} gating and adaptively learnable for each layers. This design aims to facilitate the model's ability to absorb image conditions in a layer-dependent manner. As the intermediate layers of the U-Net are more associated with object shapes or poses, and the two-end layers are more linked to appearance~\cite{voynov2023p}, we expect that the image features will primarily influence the videos' appearance while exerting relatively less impact on the shape.

\begin{figure}[!t]
    \centering
    \includegraphics[width=1\linewidth]{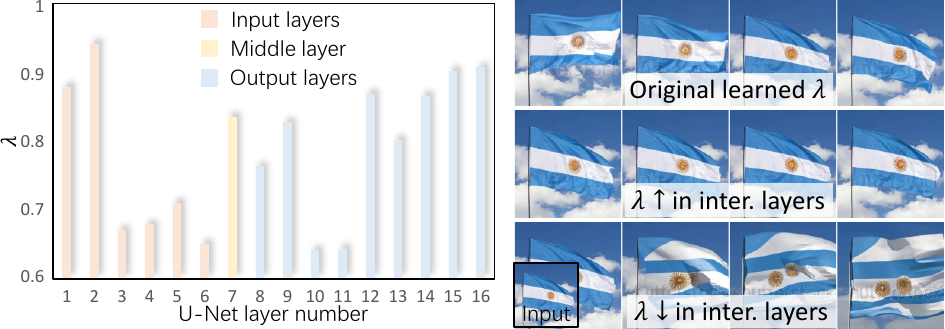}
    \vspace{-5mm}
    \caption{Visualization of the learned $\lambda$ across U-Net layers (left), and visual comparisons when manually adjusting $\lambda$ (right).}
    \label{fig:lambda}
    \vspace{-4mm}
\end{figure}

\vspace{-4mm}
\paragraph{Observations and analysis of $\lambda$.} Figure~\ref{fig:lambda} (left) illustrates the learned coefficients across different layers, indicating that the image information has a more significant impact on the two-end layers w.r.t. the intermediate layers. To explore further, we manually alter $\lambda$ in the intermediate layers. As depicted in Figure~\ref{fig:lambda} (right), increasing $\lambda$ leads to suppressed cross-frame movements, while decreasing $\lambda$ poses challenges in preserving the object's shape. This observation not only align with our expectations, but also suggests that in image-conditioned diffusion models, rich-context information influences certain intermediate layers (\eg, layers 7-9) of the U-Net, enabling the model to maintain object shape similar to the input in the presence of motions.

\begin{figure}[!t]
    \centering
    \includegraphics[width=1\linewidth]{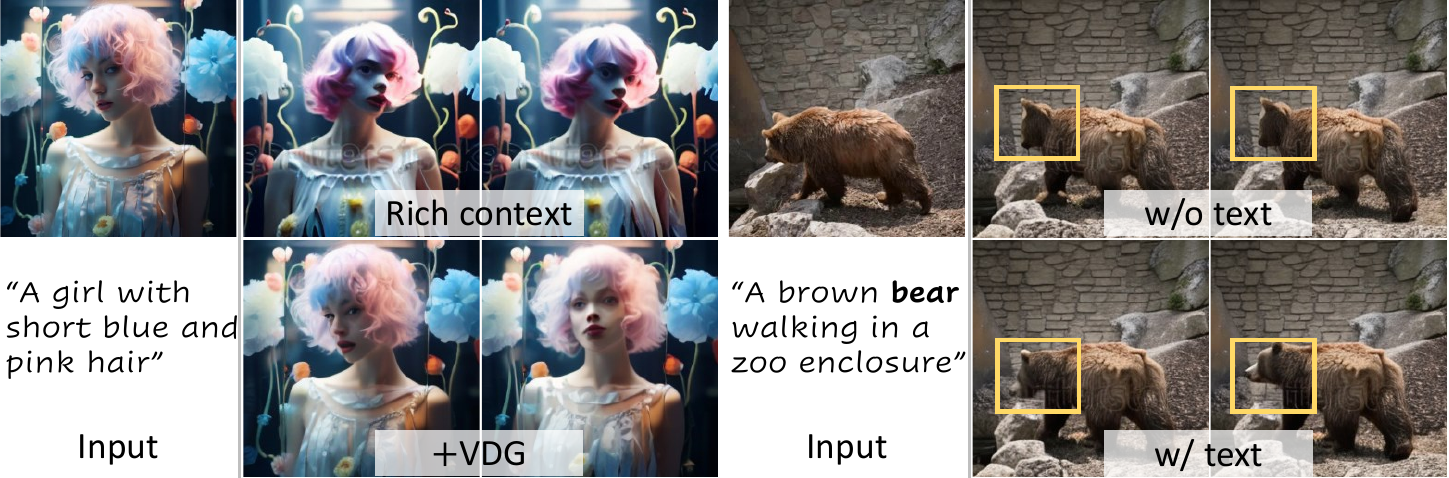}
    \vspace{-5mm}
    \caption{(Left) Comparison of animations produced using rich context representation solely, and additionally visual detail guidance (VDG). (Right) Impact of text with context representation.}
    \label{fig:visual_detail_guidance}
    \vspace{-5mm}
\end{figure}

\vspace{-4mm}
\paragraph{Visual detail guidance (VDG).}
The rich-informative context representation enables the video diffusion model to produce videos that closely resemble the input image. However, as shown in Figure~\ref{fig:visual_detail_guidance}, minor discrepancies may still occur. This is mainly due to the pre-trained CLIP image encoder's limited capability to fully preserve input image information, as it is designed to align visual and language features.
To enhance visual conformity, we propose providing the video model with additional visual details from the image. Specifically, we concatenate the conditional image with per-frame initial noise and feed them to the denoising U-Net as a form of guidance. Therefore, in our proposed dual-stream image injection paradigm, the video diffusion model integrates both global context and local details from the input image in a complementary fashion.

\vspace{-4mm}
\paragraph{Discussion.}
(i) \textit{Why are text prompts necessary when a more informative context representation is provided?}
Although we construct a text-aligned context representation, it carries more extensive information than text embedding, which may overburden the T2V model to digest them properly, \eg, causing shape distortion. Additional text prompts can offer a native global context that enables the model to efficiently utilize image information. Figure~\ref{fig:visual_detail_guidance} (right) demonstrates how incorporating text can address the issue of shape distortion in the bear's head.
Furthermore, as a still image typically contains multiple potential dynamic variations, text prompts can effectively guide the generation of dynamic content tailored to user preferences (see Sec.~\ref{sec:discussion}).
(ii) \textit{Why is a rich context representation necessary when the visual guidance provides the complete image?}
As previously mentioned, the pre-trained T2V model comprises a semantic control space (text embedding) and a complementary random space (initial noise).
While the random space effectively integrates low-level information, concatenating the noise of each frame with a fixed image induces spatial misalignment potentially, which may misguide the model in uncontrollable directions. Regarding this, the precise visual context supplied by the image embedding can assist in the reliable utilization of visual details. The corresponding ablation study is presented in Sec.~\ref{subsec:ablation}.

\subsection{Training Paradigm}
\label{subsec:training}
The conditional image is integrated through two complementary streams, which play roles in context control and detail guidance, respectively. To modulate them in a cooperative manner, we device a dedicated training strategy consisting of three stages, \ie, (i) training the image context representation network $\mathcal{P}$, (ii) adapting $\mathcal{P}$ to the T2V model, and (iii) joint fine-tuning with VDG.

Specifically, to offer the image information to the T2V model in a compatible fashion, we propose to train a context representation network $\mathcal{P}$ to extract text-aligned visual information from the input image. Considering the fact that $\mathcal{P}$ takes numerous optimization steps to converge, 
we propose to train it based on a lightweight T2I model instead of a T2V model, allowing it to focus on image context learning, and then adapt it to the T2V model by jointly training $\mathcal{P}$ and spatial layers (in contrast to temporal layers) of the T2V model.
After establishing a compatible context conditioning branch for T2V, we concatenate the input image with per-frame noise for joint fine-tuning to enhance visual conformity. Here we only fine-tune $\mathcal{P}$ and the VDM's spatial layers to avoid disrupting the pre-trained T2V model's temporal prior knowledge with dense image concatenation, which could lead to significant performance degradation and contradict our original intention. Additionally, we randomly select a video frame as the image condition based on two considerations: (i) to prevent the network from learning a shortcut that maps the concatenated image to a frame in the specific location, and (ii) to force the context representation to be more flexible to avoid offering the over-rigid information for a specific frame, \ie, the objective in the context learning based on T2I.

\begin{figure*}[!ht]
    \centering
    \includegraphics[width=1\linewidth]{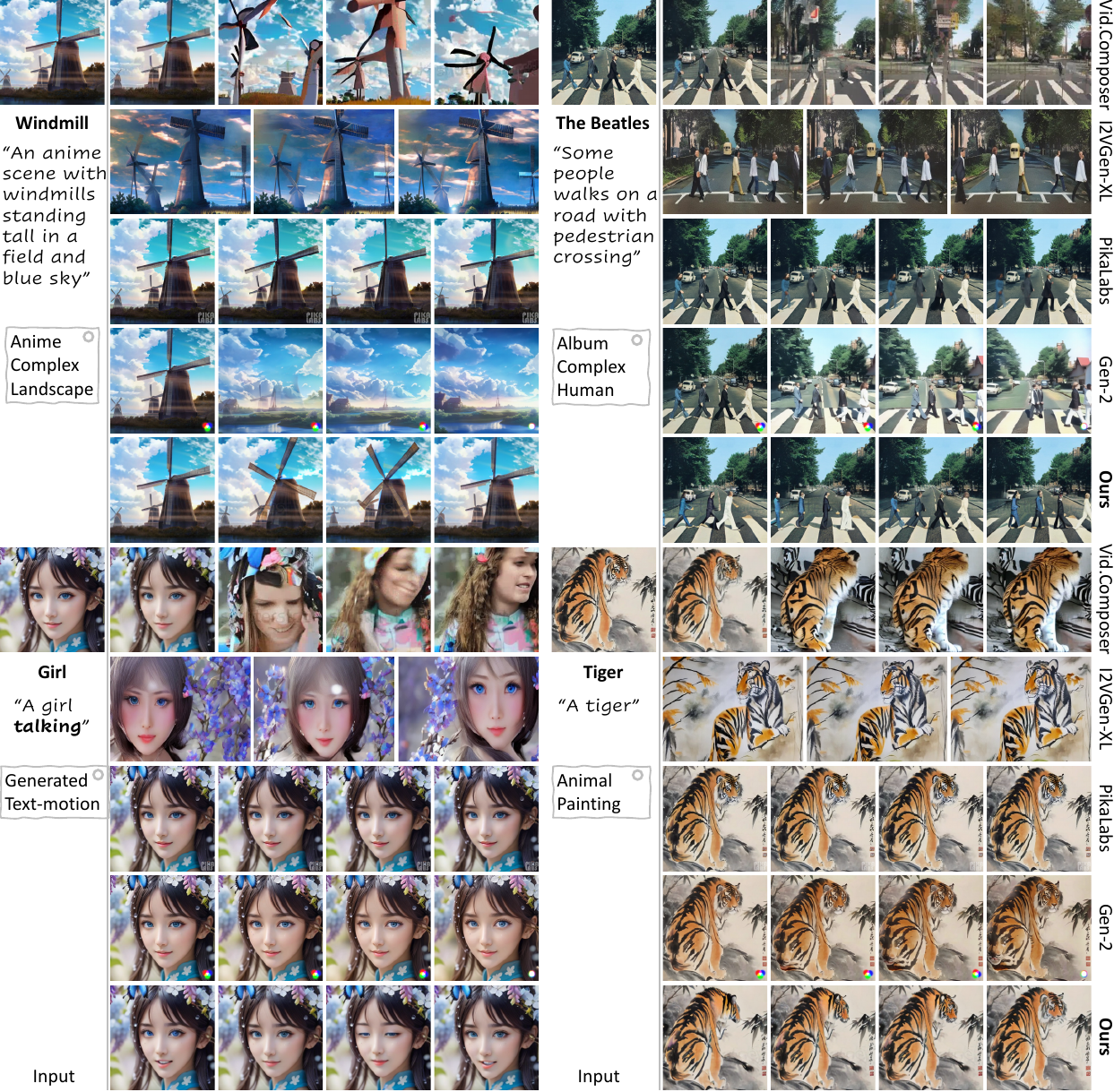}
    \vspace{-6mm}
    \caption{Visual comparisons of image animation results from VideoComposer, I2VGen-XL, PikaLabs, Gen-2, and our DynamiCrafter.}
    \label{fig:qualitative}
    \vspace{-5mm}
\end{figure*}

%% file: tex/4_experiment.tex
\section{Experiment}
\label{sec:experiment}

\subsection{Implementation Details}
Our development is based on the open-source T2V model VideoCrafter~\cite{chen2023videocrafter1} (@$256\times 256$ resolution) and T2I model Stable-Diffusion-v2.1 (SD)~\cite{rombach2022high}. We firstly train $\mathcal{P}$ and the newly injected image cross-attention layers based on SD, with 1000K steps on the learning rate $1\times 10^{-4}$ and valid mini-batch size 64. Then we replace SD with VideoCrafter and further fine-tune $\mathcal{P}$ and spatial layers with 30K steps for adaptation, and additional 100K steps with image concatenation on the learning rate $5\times 10^{-5}$ and valid mini-batch size 64. Our DynamiCrafter was trained on WebVid-10M~\cite{bain2021frozen} dataset by sampling 16 frames with dynamic FPS at the resolution of $256\times 256$ in a batch. At inference, we adopt DDIM sampler~\cite{song2021denoising} with multi-condition classifier-free guidance~\cite{ho2022classifier}. Specifically, similar to video editing~\cite{esser2023structure}, we introduce two guidance scales $s_\text{img}$ and $s_\text{txt}$ to text-conditioned image animation, which can be adjusted to trade off the impact of two control signals:
\vspace{-3mm}
\begin{align}
    \nonumber \mathbf{\hat{\epsilon}}_{\theta}\left(\mathbf{z}_t, \mathbf{c}_\text{img}, \mathbf{c}_\text{txt}\right) &=   \mathbf{\epsilon}_{\theta}\left(\mathbf{z}_t, \varnothing, \varnothing\right)\\ \nonumber & +s_\text{img}(\mathbf{\epsilon}_{\theta}\left(\mathbf{z}_t, \mathbf{c}_\text{img}, \varnothing\right)-\mathbf{\epsilon}_{\theta}\left(\mathbf{z}_t, \varnothing, \varnothing\right))\\ \nonumber & +s_\text{txt}(\mathbf{\epsilon}_{\theta}\left(\mathbf{z}_t, \mathbf{c}_\text{img}, \mathbf{c}_\text{txt}\right)-\mathbf{\epsilon}_{\theta}\left(\mathbf{z}_t, \mathbf{c}_\text{img}, \varnothing\right)). 
\end{align}
\vspace{-8mm}

\subsection{Quantitative Evaluation}
\paragraph{Metrics and datasets.} To evaluate the quality and temporal coherence of synthesized videos in both the spatial and temporal domains, we report Fr\'echet Video Distance (FVD)~\cite{unterthiner2019fvd} as well as Kernel Video Distance (KVD)~\cite{unterthiner2019fvd}.
Following~\cite{zhou2022magicvideo,blattmann2023align}, we evaluate the zero-shot generation performance of all the methods on UCF-101~\cite{soomro2012ucf101} and MSR-VTT~\cite{xu2016msr}.  
To further investigate the perceptual conformity between the input image and the animation results, we introduce Perceptual Input Conformity (PIC), which is computed by $\frac{1}{L}\sum_l (1-D(\mathbf{x}^\text{in},\mathbf{x}^l))$, where $\mathbf{x}^\text{in},\mathbf{x}^l$, $L$ are the input image, video frames, and video length, respectively, and we adopt the perceptual distance metric DreamSim~\cite{fu2023dreamsim} as the distance function $D(\cdot,\cdot)$.
We evaluate each error metric at the resolution of $256\times 256$ with 16 frames.

As open-domain image animation is a nascent area of computer vision, there are limited publicly available research works for comparison. We evaluate our method against VideoComposer~\cite{wang2023videocomposer} and I2VGen-XL~\cite{I2VGen-XL}, with the quantitative results presented in Table~\ref{tab:i2v_quan}. According to the results, our proposed method significantly outperforms previous approaches in all evaluation metrics, except for KVD on UCF-101, thanks to the effective dual-stream image injection design for fully exploiting the video diffusion prior.

\begin{table}[t]
  \caption{Quantitative comparisons with state-of-the-art open-domain image-to-video generation methods on UCF-101 and MSR-VTT for the zero-shot setting.}
  \label{tab:i2v_quan}
  \vspace{-0.8em}
\resizebox{\linewidth}{!}{
  \setlength{\tabcolsep}{2.7pt}
  \centering
  \begin{tabular}{lcccccc}
  \toprule
    \multirow{2}{*}{Method}  & \multicolumn{3}{c}{UCF-101}  & \multicolumn{3}{c}{MSR-VTT} \\
    \cmidrule(lr){2-4}\cmidrule(lr){5-7}
     &FVD $\downarrow$ & KVD $\downarrow$ &PIC $\uparrow$ &FVD $\downarrow$ & KVD $\downarrow$ &PIC $\uparrow$\\
    \midrule
    VideoComposer &576.81 &65.56 &0.5269 &377.29 &26.34 &0.4460\\ 
    I2VGen-XL    &571.11 &\best{58.59} &0.5313 &289.10 &14.70 &0.5352\\
    \rowcolor{Gray}
    Ours          &\best{429.23} &62.47 &\best{0.6078} &\best{234.66} &\best{13.74} &\best{0.5803}\\
  \bottomrule
  \end{tabular}
}
\vspace{-4mm}
\end{table}

\subsection{Qualitative Evaluation}
 In addition to the aforementioned approaches, we include two more proprietary commercial products, \ie, PikaLabs~\cite{PikaLabs} and Gen-2~\cite{Gen-2}, for qualitative comparison. Note that the results we accessed on Nov. 1$^{\text{st}}$,~2023 might differ from the current product version due to rapid version iterations. Figure~\ref{fig:qualitative} presents the visual comparison of image animation results with various content and styles. Among all compared methods, our approach generates temporally coherent videos that adhere to the input image condition. In contrast, VideoComposer struggles to produce consistent video frames, as subsequent frames tend to deviate from the initial frame due to inadequate semantic understanding of the input image. I2VGen-XL can generate videos that semantically resemble the input images but fails to preserve intricate local visual details and produce aesthetically appealing results. As commercial products, PikaLabs and Gen-2 can produce appealing high-resolution and long-duration videos. However, Gen-2 suffers from sudden content changes (the `Windmill' case) and content drifting issues (`The Beatles' and `Girl' cases). PikaLabs tends to generate still videos with less dynamic and exhibits blurriness when attempting to produce larger dynamics (`The Beatles' case). It is worth noting that our method allows dynamic control through text prompts while other methods suffers from neglecting the text modality (\eg, \emph{talking} in the `Girl' case). More videos are provided in the \emph{Supplement}.

\begin{table}[t]
  \caption{User study statistics of the preference rate for Motion Quality (M.Q.) \& Temporal Coherence (T.C.), and selection rate for visual conformity to the input image (I.C.=Input Conformity).}
  \vspace{-0.8em}
  \label{tab:user_study}
\resizebox{\linewidth}{!}{
  \setlength{\tabcolsep}{2.1pt}
  \centering
  \begin{tabular}{lcccca}
  \toprule
    \multirow{2}{*}{Property} &\multicolumn{2}{c}{Proprietary} & \multicolumn{3}{c}{Open-source} \\
    \cmidrule(lr){2-3}\cmidrule(lr){4-6}
    & PikaLabs&Gen-2& VideoComposer & I2VGen-XL   & \mc{1}{\multirow{1}{*}{Ours}} \\
    \midrule
    M.Q. $\uparrow$& 28.60\%& 22.91\%&  2.09\%&  7.56\%& \best{38.84}\%\\
    T.C. $\uparrow$& 32.09\%& 26.05\%&  2.21\%&  6.51\%& \best{33.14}\%\\
    I.C. $\uparrow$&  79.07\%& 64.77\%& 18.14\%& 15.00\%& \best{79.88}\%\\
  \bottomrule
  \end{tabular}
}
\vspace{-4mm}
\end{table}
\vspace{-4mm}
\paragraph{User study.}
We conduct a user study to evaluate the perceptual quality of the generated images. 
The participants are asked to choose the best result in terms of motion quality and temporal coherence, and to select the results with good visual conformity to the input image for each case. The statistics from 49 participants' responses are presented in Table~\ref{tab:user_study}. Our method demonstrates significant superiority over other open-source methods. Moreover, our method achieves comparable performance in terms of temporal coherence and input conformity compared to commercial products, while exhibiting superior motion quality.

\subsection{Ablation Studies}
\label{subsec:ablation}
\begin{figure}[!t]
    \centering
    \includegraphics[width=1\linewidth]{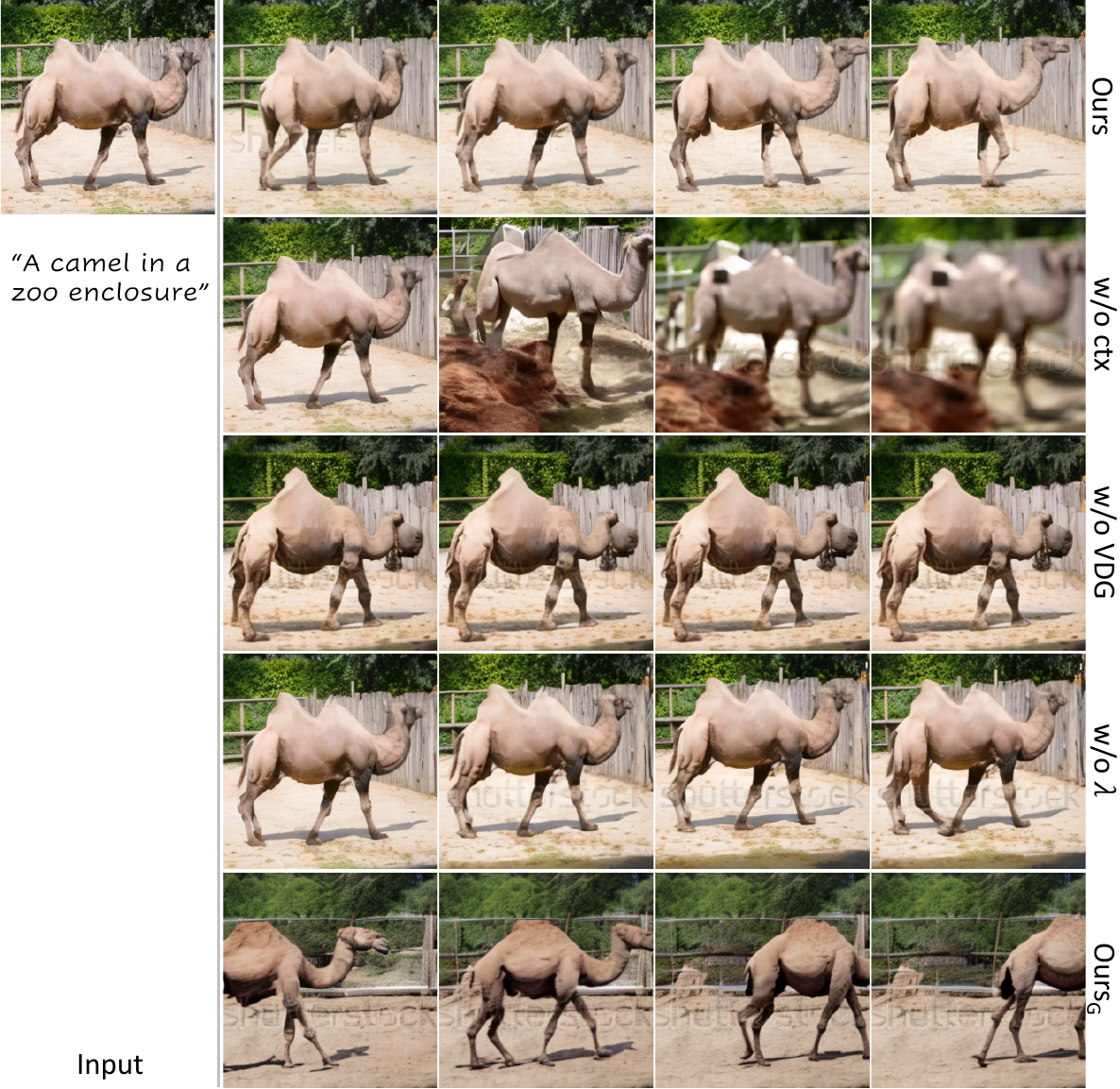}
    \vspace{-6mm}
    \caption{Visual comparisons of different variants of our method.}
    \label{fig:ablation1}
    \vspace{-5mm}
\end{figure}

\paragraph{Dual-stream image injection.} To investigate the roles of each image conditioning stream, we examine two variants: \textbf{i). Ours w/o ctx}, by removing the context conditioning stream, \textbf{ii). Ours w/o VDG}, by removing the visual detail guidance stream. Table~\ref{tab:ablation_1} presents a quantitative comparison between our full method and these variants. 
The performance of `w/o ctx' declines significantly due to its inability to semantically comprehend the input image without injection of rich-context representation, leading to temporal inconsistencies in the generated videos (see the 2nd row in Figure~\ref{fig:ablation1}).
Although removing the VDG (w/o VDG) can yield better FVD scores, it causes severe shape distortions and exhibits limited motion magnitude, as the remaining context condition can only provide semantic-level image information.
Moreover, while it achieves a higher PIC score, it fails to capture all the visual details of the input image, as evidenced by the 3rd row in Figure~\ref{fig:ablation1}.
\begin{table}[t]
  \caption{Ablation study on the dual-stream image injection and training paradigm.}
  \label{tab:ablation_1}
  \vspace{-0.8em}
\resizebox{\linewidth}{!}{
  \setlength{\tabcolsep}{2.1pt}
  \centering
  \begin{tabular}{lacccccc}
  \toprule
    \multirow{2}{*}{Metric} & \mc{1}{\multirow{2}{*}{Ours}} &\multicolumn{4}{c}{Dual-stream image injection} & \multicolumn{2}{c}{Training paradigm}\\
    \cmidrule(lr){3-6}\cmidrule(lr){7-8}
    & \mc{1}{}& w/o ctx & w/o VDG& w/o $\lambda$ & Ours$_\text{G}$   & Ft. ent. & 1st frame \\
    \midrule
    FVD $\downarrow$ & 234.66& 372.80 & 159.24 &  241.38  &  286.84  & 364.11& 309.23\\
    PIC $\uparrow$ & 0.5803& 0.4916   & 0.6945 &  0.5708  &  0.5717  &0.5564 &0.5673\\
  \bottomrule
  \end{tabular}
}
\vspace{-4mm}
\end{table}

We then study several key designs in the context representation stream: adaptive gating $\lambda$ and full visual tokens in CLIP image encoder.
Eliminating the adaptive gating $\lambda$ (w/o $\lambda$) leads to a slight decrease in model performance. This is because, without considering the nature of the denoising U-Net layers, context information cannot be adaptively integrated into the T2V model, resulting in shaky generated videos and unnatural motions (see the 4th row in Figure~\ref{fig:ablation1}). 
On the other hand, using a strategy (Ours$_\text{G}$) like I2VGen-XL that utilizes a single CLIP global token may generate results that are only semantically similar to the input due to the absence of full image extent. 
In contrast, our full method effectively leverages the video diffusion prior for image animation with natural motion, coherent frames, and visual conformity to the input image.

\begin{figure}[!t]
    \centering
    \includegraphics[width=1\linewidth]{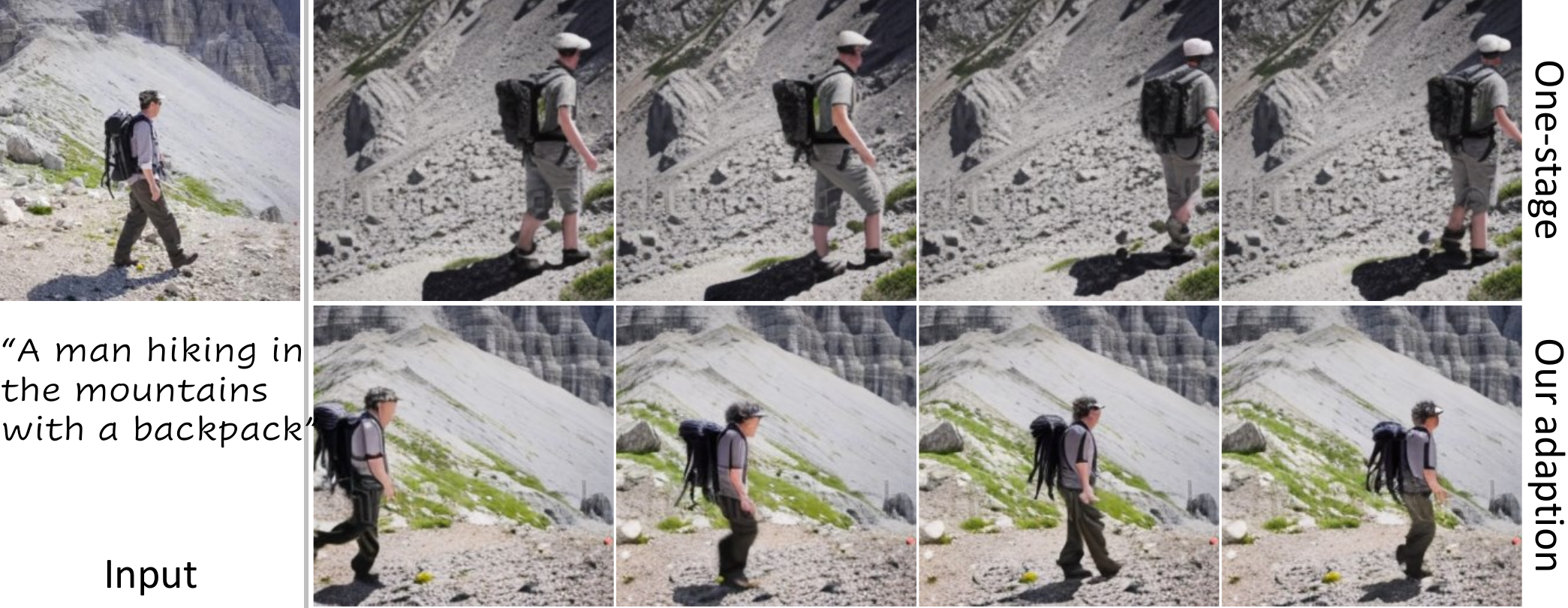}
    \vspace{-6mm}
    \caption{Visual comparisons of the context conditioning stream learned in one-stage and our two-stage adaption strategy.}
    \label{fig:ablation2.1}
    \vspace{-3mm}
\end{figure}

\begin{figure}[!t]
    \centering
    \includegraphics[width=1\linewidth]{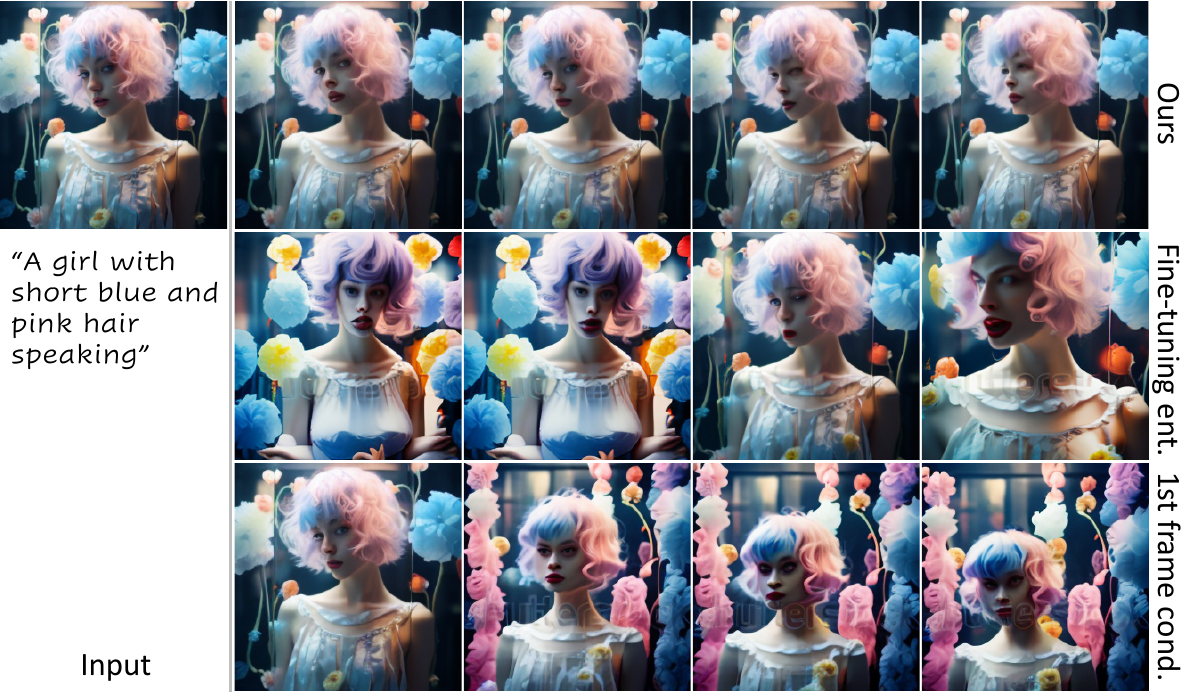}
    \vspace{-6mm}
    \caption{Visual comparisons of different training paradigms.}
    \label{fig:ablation2.2}
    \vspace{-4.5mm}
\end{figure}

\vspace{-4.5mm}
\paragraph{Training paradigm.}
We further examine the specialized training paradigm to ensure the model works as expectation. We firstly construct a baseline by training the context representation network $\mathcal{P}$ based on the pre-trained T2V and keeping other settings unchanged. As illustrated in Figure~\ref{fig:ablation2.1}, this baseline (one-stage) converges at a significantly slow pace, resulting in only coarse-grained context conditioning with the same optimization steps. This may potentially make it challenging for the T2V model to harmonize the dual-stream conditions after incorporating the VDG.

After obtaining a compatible context conditioning stream $\mathcal{P}$, we further incorporate image concatenation with per-frame noise to enhance visual conformity by jointly fine-tuning $\mathcal{P}$ and \emph{spatial layers} of the T2V model. We construct a baseline by fine-tuning the entire T2V model, and the quantitative comparison in Table~\ref{tab:ablation_1} (Ft. ent.) shows that this baseline results in an unstable model that is prone to collapse, disrupting the temporal prior. Additionally, to study the effectiveness of our random selection conditioning strategy, we train a baseline (1st frame cond.) that consistently uses the first video frame as the conditional image. Table~\ref{tab:ablation_1} reveals its inferior performance in terms of both FVD and PIC, which can be attributed to the ``content sudden change" effect observed in the generated videos (Figure~\ref{fig:ablation2.2} (bottom)). We hypothesize that the model may discover a suboptimal shortcut for mapping the concatenated image to the first frame while neglecting other frames.

%% file: tex/5_discussion.tex
\section{Discussions on Motion Control using Text}
\label{sec:discussion}

\begin{figure}[!t]
    \centering
    \includegraphics[width=1\linewidth]{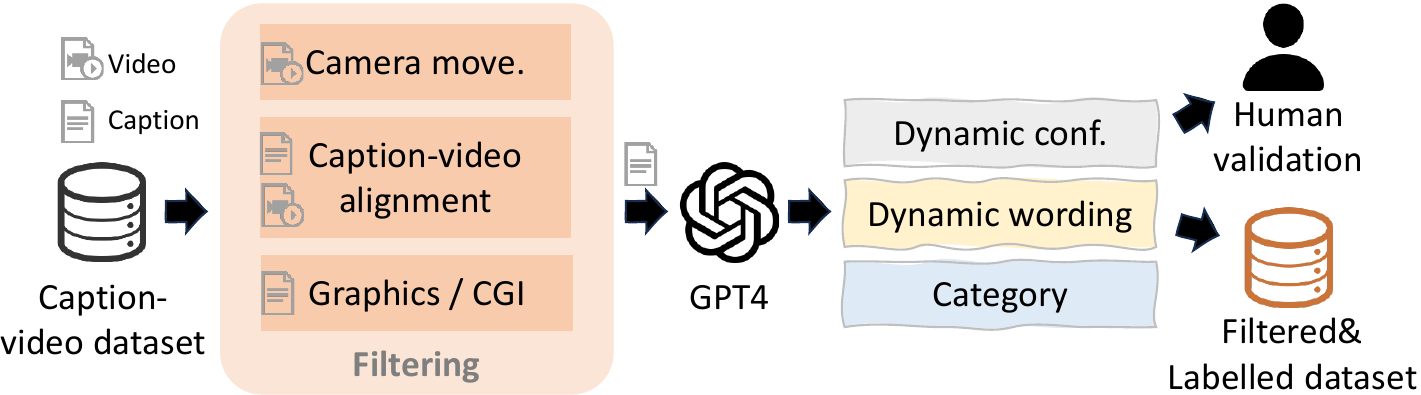}
    \vspace{-6mm}
    \caption{Illustration of dataset filtering and annotation process.}
    \label{fig:dataset}
    \vspace{-2mm}
\end{figure}
\begin{figure}[!t]
    \centering
    \includegraphics[width=1\linewidth]{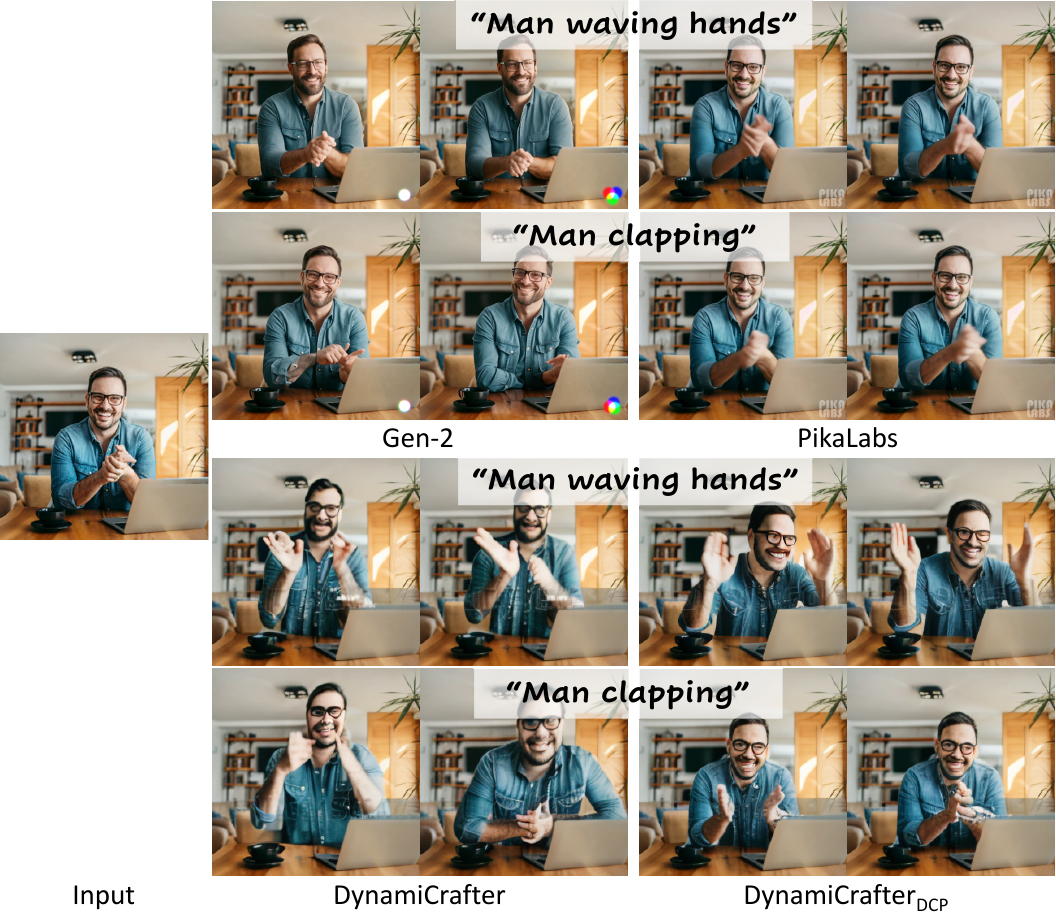}
    \vspace{-6mm}
    \caption{Visual comparisons of image animation results from different methods with motion control using text.}
    \label{fig:text_motion}
    \vspace{-4mm}
\end{figure}

Since images are typically associated with multiple potential dynamics in its context, text can complementarily guide the generation of dynamic content tailored to user preference. However, captions in existing large-scale datasets often consist of a combination of a large number of scene descriptive words and less dynamic/motion descriptions, potentially causing the model to overlook dynamics/motions during learning. For image animation, the scene description is already included in the image condition, while the motion description should be treated as text condition to train the model in a \emph{decoupled} manner, providing the model with stronger text-based control over dynamics.

\vspace{-4mm}
\paragraph{Dataset construction.}
To enable the decoupled training, we construct a dataset by filtering and re-annotating the WebVid10M dataset, as illustrated in Figure~\ref{fig:dataset}. The constructed dataset contains captions with purer \emph{dynamic wording}, such as ``Man doing push-ups.", and categories, \eg., \texttt{human}.

We then train a model DynamiCrater$_\text{DCP}$ using the dataset and validate its effectiveness with 40 image-prompt testing cases featuring human figures with ambiguous potential actions, and prompts describing various motions (\eg., ``Man waving hands" and ``Man clapping"). We measure the average CLIP similarity (CLIP-SIM) between the prompt and video results, and DynamiCrater$_\text{DCP}$ improves the performance from 0.17 to 0.19 in terms of CLIP-SIM score. The visual comparison in Figure~\ref{fig:text_motion} shows that Gen-2 and PikaLabs cannot support motion control using text, while our DynamiCrafter reflects the text prompt and is further enhanced in DynamiCrafter$_\text{DCP}$ with the proposed decoupled training. More details are in the \emph{Supplement}.

%% file: tex/6_application.tex
\section{Applications}
\label{sec:application}

\begin{figure}[!t]
    \centering
    \includegraphics[width=1\linewidth]{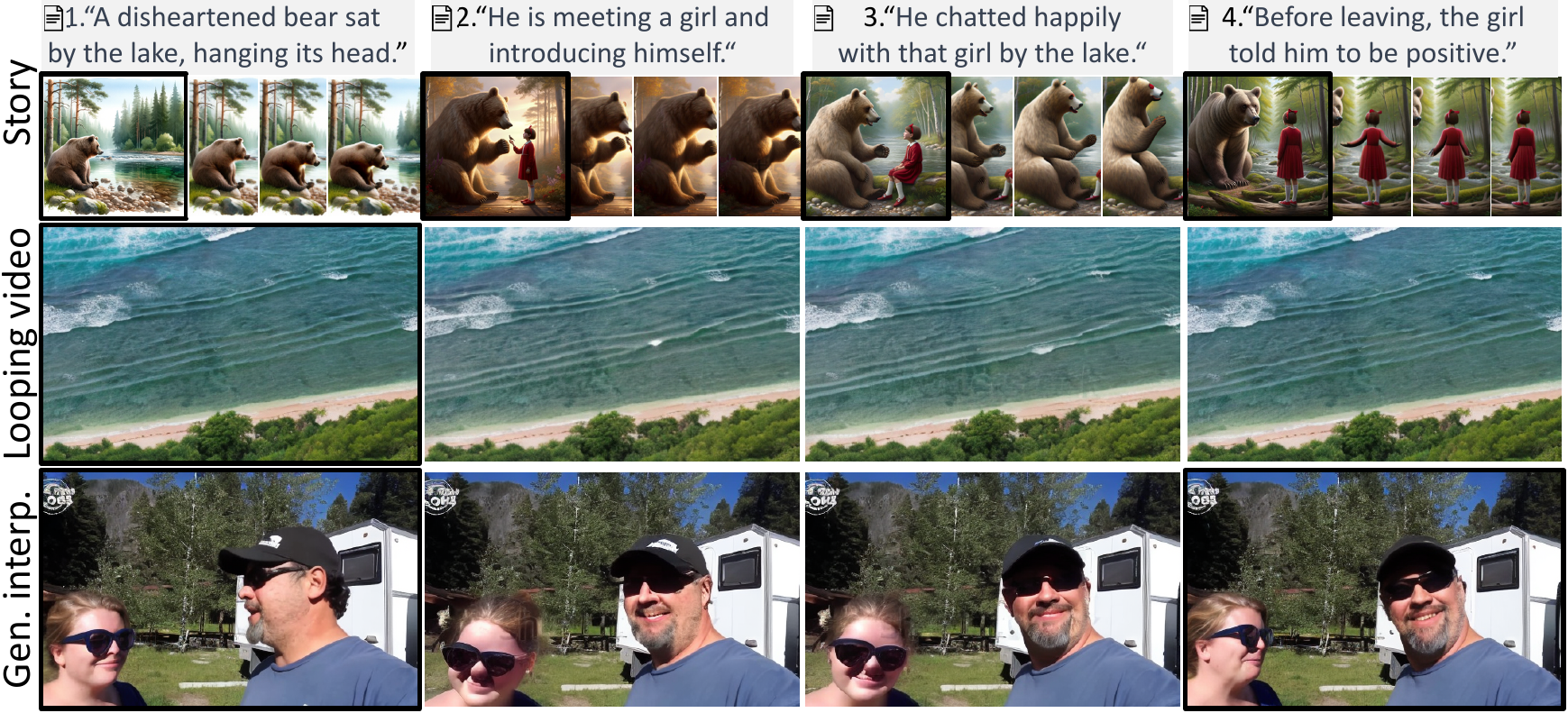}
    \vspace{-5.5mm}
    \caption{Applications of our \emph{DynamiCrafter}. $\square$: input images.}
    \label{fig:application}
    \vspace{-4mm}
\end{figure}

\emph{DynamiCrafter} can be easily adapted to support additional applications. \textbf{i). Storytelling with shots.}
First, we utilize ChatGPT (equipped with DALL-E~3~\cite{shi2020improving}) to generate a story script and corresponding shots (images).  And then storytelling videos can be generated by animating those shots with story scripts using DynamiCrafter, as displayed in Figure~\ref{fig:application} (top).
\textbf{ii). Looping video generation.}
With minor modifications, our framework can be adapted to facilitate the generation of looping videos. Specifically, we provide both $\mathbf{x}^1$ and $\mathbf{x}^{L}$ as visual detail guidance and leave other frames as empty during training. During inference, we set both of them as the input image. Additionally, we experiment with building this application on top of a higher-resolution ($320\times 512$) version of VideoCrafter. The looping video result is shown in Figure~\ref{fig:application} (middle). 
\textbf{iii). Generative frame interpolation.}
Furthermore, the modified model enables generative frame interpolation by set the input images $\mathbf{x}^1$ and $\mathbf{x}^{L}$ differently, as shown in Figure~\ref{fig:application} (bottom).

%% file: tex/8_conclusion.tex
\section{Conclusion}
In this study, we introduced \emph{DynamiCrafter}, an effective framework for animating open-domain images by leveraging pre-trained video diffusion priors with the proposed dual-stream image injection mechanism and dedicated training paradigm. Our experimental results highlight the effectiveness and superiority of our approach compared to existing methods. 
Furthermore, we explored text-based dynamic control for image animation with the constructed dataset. Lastly, we demonstrated the versatility of our framework across various applications and scenarios.

%% file: X_suppl.tex
\clearpage
\newcommand{\nocontentsline}[3]{}
\newcommand{\tocless}[2]{\bgroup\let\addcontentsline=\nocontentsline#1{#2}\egroup}

\newcommand{\Appendix}[1]{
  \refstepcounter{section}
  \section*{Appendix \thesection:\hspace*{1.5ex} #1}
  \addcontentsline{toc}{section}{Appendix \thesection}
}
\newcommand{\SubAppendix}[1]{\tocless\subsection{#1}}
\maketitlesupplementary
\appendix

\begin{table*}[!ht]
  \caption{Summary of open-domain (text-)image-to-video generation methods. $^*$The resolution is obtained by inputting a square-sized image into these methods.}
  \label{tab:baselines}
\resizebox{\linewidth}{!}{
  \centering
  \begin{tabular}{lccccccc}
  \toprule
    Method &Open-source &Verison (Date) & Resolution$^*$ & Duration & FPS & Text input & Description (visual condition injection) \\
    \midrule
    \multirow{2}{*}{VideoComposer}  & \multirow{2}{*}{\cmark} &\multirow{2}{*}{23.06.29}& \multirow{2}{*}{$256\times 256$} & \multirow{2}{*}{2s} & \multirow{2}{*}{8} & \multirow{2}{*}{\cmark} & The encoded image information is injected via \\
    & & & & & & &  frame-wise concatenation with the noisy latent.\\
    \multirow{2}{*}{I2VGen-XL} & \multirow{2}{*}{\cmark} &\multirow{2}{*}{23.10.30}& \multirow{2}{*}{$256\times 448$} & \multirow{2}{*}{3s} & \multirow{2}{*}{8} & \multirow{2}{*}{\xmark} & Image information is injected by cross-attention via  \\
    & & & & & & &  the global token from CLIP image encoder.\\
    PikaLabs & \xmark &23.11.01& $768\times 768$ & 3s & 24 & \cmark & Unknown\\
    Gen-2 & \xmark &23.11.01& $896\times 896$ & 4s & 24 & \cmark & Unknown\\
  \bottomrule
  \end{tabular}
}
\end{table*}
\input{supp/hyper-parameters}

\tableofcontents
\addtocontents{toc}{\setcounter{tocdepth}{2}}

\vspace{4mm}
Please check our project page \url{https://doubiiu.github.io/projects/DynamiCrafter} for video results.

\section{Implementation Details}
\label{sec:implmentation}

\subsection{Network Architecture}
\begin{figure}[!t]
    \centering
    \includegraphics[width=1\linewidth]{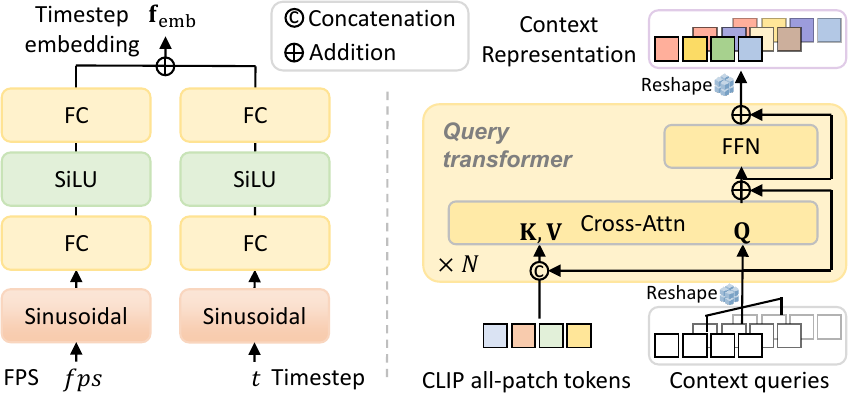}
    \caption{Network architecture of the FPS embedding layer (left) and query transformer for context representation learning (right).}
    \label{fig:network_arch}
\end{figure}

Our \emph{DynamiCrafter} is built upon VideoCrafter, a latent VDM-based text-to-video (2TV) generation model, so we recommend that readers refer to VideoCrafter~\cite{chen2023videocrafter1} for more details of the T2V backbone. It is worth noting that our approach of leveraging the video diffusion prior for image animation can theoretically be applied to any other T2V diffusion models that incorporate a cross-attention text-conditioning mechanism.
To improve the reproducibility of our method, we provide a more detailed description of the network architecture for the FPS embedding layer and context query transformer.
As depicted in Figure~\ref{fig:network_arch} (left), the FPS condition is embedded via Sinusoidal and several Fully-Connected (FC) layers activated by SiLU~\cite{hendrycks2016gaussian}, which is then added to the timestep embedding $\mathbf{f}_\text{emb}$. In Figure~\ref{fig:network_arch} (right), the context query transformer first projects the concatenation of frame-wise context queries and CLIP tokens into \emph{keys} and \emph{values}, while projecting context queries solely into \emph{queries}. The cross-attention results are subsequently computed using the keys, values, and queries, and projected via a Feed-forward layer. The final frame-wise context representation is then employed through the spatial dual-attn transformer in the denoising U-Net, as illustrated in Figure 1 and Equation 2 in the main paper.

\subsection{Hyper-parameters}
Following~\cite{blattmann2023align}, all architecture parameter details, diffusion process details, as well as training hyper-parameters are provided in Table~\ref{table:hyperparameters}, which should be mostly self-explanatory. Here we give some additional description for some parameters:
\begin{itemize}
    \item Input channels (Architecture): The number of input tensor channels for the denoising U-Net, which is twice the channel number of $\mathbf{z}_t$ due to the channel-wise concatenation of visual detail guidance.
    \item CA ctx sequence length (Dual-CA Conditioning): The token length of the context representations for each frame.
\end{itemize}

\subsection{Training}
Since we concatenate the conditional image latent with noisy latents in the channel dimension (\ie, visual detail guidance in Section 3.2 of the main paper), we add additional input channels to the first convoluional layer. All available weights of the video diffusion model are initialized from the pre-trained checkpoints, and weights that operate on the newly added input channels are initialized to zero.
We utilize only eight NVIDIA V100 GPUs to fine-tune the T2V model that is relatively resource-friendly in the context of developing an image-to-video diffusion model.
As mentioned in Section 4.1 of the main paper, we fine-tune the T2V model using WebVid10M, primarily consists of real-world videos. Despite this, the model demonstrates strong generalizability when animating images that are even outside its domain, such as anime or paintings.

\section{Additional Evaluation Details}
\label{sec:evaluation}
\subsection{Dataset and metric}
To evaluate the quality and temporal coherence of synthesized videos in both the spatial and temporal domains, we report Fr\'echet Video Distance (FVD)~\cite{unterthiner2019fvd} as well as Kernel Video Distance (KVD)~\cite{unterthiner2019fvd}, which evaluate video quality by measuring the feature-level similarity between synthesized and real videos based on the Fr\'echet distance and kernel methods, respectively. 
Specifically, they are computed by comparing 2048 model samples with samples from evaluation datasets, where we adopt commonly used UCF-101~\cite{soomro2012ucf101} and MSR-VTT~\cite{xu2016msr} for benchmarking. For UCF-101, we directly use UCF class names~\cite{blattmann2023align} as text conditioning, while for MSR-VTT, we utilize accompanied captions of each video from the dataset. 
We evaluate each error metric at the resolution of $256\times 256$ with 16 frames.

\subsection{Baselines}
In the emerging field of open-domain image animation, there are limited baselines available for comparison. In this study, We evaluate our method against two open-source research works, \ie, VideoComposer~\cite{wang2023videocomposer} and I2VGen-XL~\cite{I2VGen-XL}, and two proprietary commercial products, \ie, PikaLabs~\cite{PikaLabs} and Gen-2~\cite{Gen-2}, which are summarized in Table~\ref{tab:baselines}. Note that we employ the image-to-video (first-stage) generation of I2VGen-XL for the evaluation experiment, as its refinement stage (text-to-video) primarily functions as a super-resolution process, with the dynamics and temporal coherence already determined by the first stage.

\section{User Study}
\label{sec:user_study}
The designed user study interface is shown in Figure~\ref{fig:user_study_screenshot}. We collect 20 image cases with a wide range of content and styles from the Internet and create corresponding captions. We then generate the image animation results by either executing the official code~\cite{wang2023videocomposer,I2VGen-XL} or accessing the online demo interface~\cite{Gen-2,PikaLabs}. For the user study, we use these video results produced by shuffled methods based on the same input still image (and text prompt, if applicable). In addition, we conceal the lower watermark region and standardize the all the produced results by first setting FPS=8, and then trimming the videos to two seconds at the same resolution level ($256\times 448$ for I2VGen-XL, while $256\times 256$ for other methods). This process ensures a fair comparison by eliminating the potential impact of engineering tricks. 

The user study is expected to be completed with 5--10 minutes (20 cases $\times$ 3 sub-questions $\times$ 5--10 seconds for each judgement). To remove the impact of random selection, we filter out those comparison results completed within three minutes. For each participant, the user study interface shows 20 video comparisons, and the participant is instructed to evaluate the videos for three times, i.e answering the following questions respectively: (i) ``Which one has the best motion/dynamic quality?"; (ii) ``Which one has the best temporal coherence?"; (iii) ``Which results conform to the input image?". Finally, we received 49 valid responses from the participants.

\section{Details of Constructed Dataset}
\label{sec:dataset}

\subsection{Dataset construction details}
As depicted in Figure 8 of the main paper, we first filter out data with large camera movement, poor caption-video alignment, and Graphics/CGI content. We then feed captions to GPT4~\cite{openai2023gpt4} (temperature=0.2, frequency\_penalty=0) to generate the following: \emph{dynamic confidence}, which represents the level of confidence that the caption describes a dynamic scene, \emph{dynamic wording}, such as ``man doing push-ups", and the category of this dynamic scene. The used dialog instructions are as follows:
\noindent\makebox[\linewidth]{\rule{\linewidth}{0.4pt}}
\vspace{-8mm}
\paragraph{User:} You are an expert assistant. There some caption-video pairs in the dataset, and you can only access the captions. You need to check if the caption describes the scene dynamics in the video, for example some actions of humans and animals, etc. 
Please output the following: 
1. Dynamic confidence. Output how confident you feel that it is describing a dynamic scene, from 0 to 100. 0 means lowest confidence and 100 means the highest confidence. 
2. Dynamic wording. Output the subject followed by actions and corresponding objects, for example ``man playing football". It must be compact. Output ``none" when the caption does not describe any scene dynamics. 
3. Dynamic source category. Classify the dynamics, the categories are \texttt{human}, \texttt{animal}, \texttt{nature}, \texttt{machine}, \texttt{others}, and \texttt{none}. \texttt{none} is used when the corresponding dynamic wording is none. \texttt{nature} indicates those dynamics related to natural phenomena, while \texttt{machine} corresponds those movements related to vehicles and technical devices. 
\begin{figure}[!t]
    \centering
    \includegraphics[width=1\linewidth]{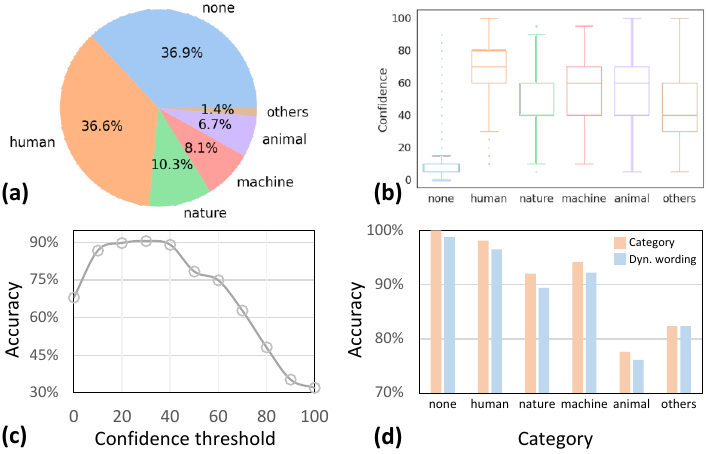}
    \caption{Statistics of the dataset and human validation results.}
    \label{fig:dataset_stat}
    \vspace{-2mm}
\end{figure}

The input is in the format of ``$\textless$question index$\textgreater$\%\%$\textless$caption$\textgreater$", The output must be in the format of ``$\textless$question index$\textgreater$\%\%$\textless$confidence$\textgreater$\%\%$\textless$dynamic wording$\textgreater$ \%\%$\textless$dynamic source category$\textgreater$". 
Here are some examples: 
\vspace{-5mm}
\paragraph{\emph{Input:}} [``1\%\%Woman in gym working out", ``2\%\%4k corporate shot of a business woman working on computer eating funny banana", ``3\%\%Rainy clouds sailing above a city", ``4\%\%View of the great salt lake", ``5\%\%Old house with a ghost in the forest at night or abandoned haunted horror house in fog."]
\vspace{-4mm}
\paragraph{\emph{Output:}} [``1\%\%80\%\%woman working out\%\%human", ``2\%\%80\%\%business woman working on computer, eating banana\%\%human", ``3\%\%50\%\%Rainy clouds sailing\%\%nature", ``4\%\%5\%\%none\%\%none", ``5\%\%10\%\%none\%\%none"].

Input captions are in an array: [caption1, caption2, $\ldots$].
\vspace{-4mm}
\paragraph{System:} Answer for every caption in the array and reply with an array of all completions.

\noindent\makebox[\linewidth]{\rule{\linewidth}{0.4pt}}

Here are some sampled inputs and outputs (w/o index):
\vspace{-5mm}
\paragraph{\emph{Input}:}[``Young man in bathrobe brushing his teeth in front of the window.", ``Summer green maple tree swinging in the wind.", ``Ripe rambutan fruits on a street market. sri lanka."]
\vspace{-5mm}
\paragraph{\emph{Output}:}[``70\%\%man brushing teeth\%\%human", ``50\%\%maple tree swinging\%\%nature", ``5\%\%none\%\%none"]

\subsection{Statistics of the dataset}
The constructed dataset contains around 2.6 million caption-video pairs, with the corresponding statistics and dynamic confidence for each category shown in Figure~\ref{fig:dataset_stat} (a) and (b), respectively. We exclude certain combinations of classes, such as `\texttt{animal}\&\texttt{human}', `\texttt{human}\&\texttt{machine}', `\texttt{animal}\&\texttt{machine}' due to their small proportions. To support potential research on motions and dynamics, we will make the annotations of the constructed dataset publicly available.


\begin{figure}[!t]
    \centering
    \includegraphics[width=1\linewidth]{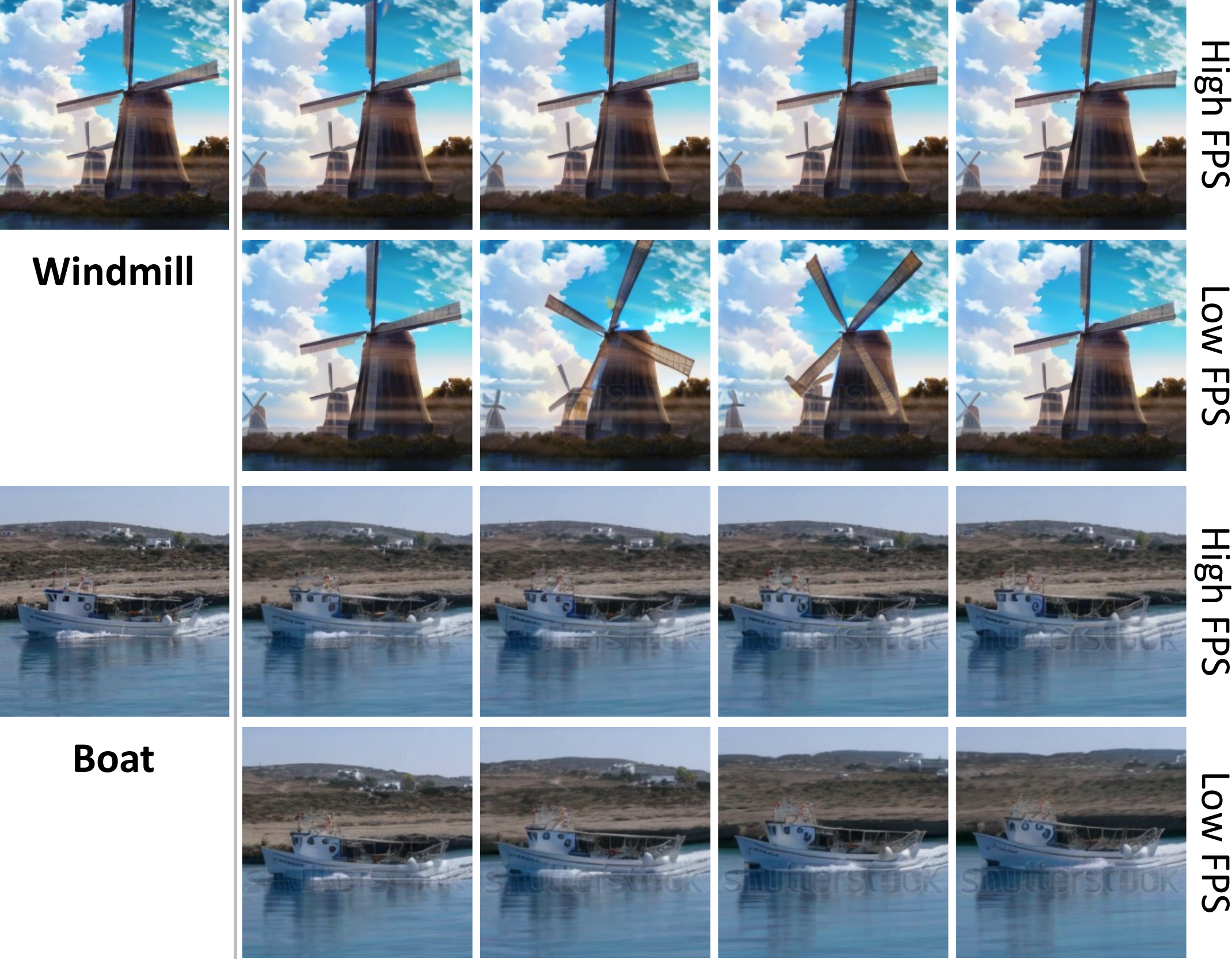}
    \caption{Visual comparisons of image animation results produced by our DynamiCrafter with FPS control.}
    \label{fig:fps_control}
\end{figure}
\subsection{Human validation on the dataset}
We also validate GPT4's responses through human judgement on randomly sampled 1K respones. We ask volunteers to determine if the original video caption describes a dynamic scene and if the dynamic wording and category generated by GPT4 are accurate. In Figure~\ref{fig:dataset_stat} (c), we plot an accuracy-threshold curve by adjusting the confidence threshold and calculating the accuracy based on human judgments of dynamic scenes. We observe that \emph{dynamic confidence}=40 serves as a sweet spot in aligning with human judgement. The accuracy of dynamic wording and category for each category are shown in Figure~\ref{fig:dataset_stat} (d). The validation results indicate that GPT4's responses generally align with human judgments, making them reliable for dataset annotation.

\subsection{DynamiCrafter$_\text{DCP}$}
Finally, we initialize DynamiCrater$_\text{DCP}$ using an intermediate checkpoint (60K iterations) from DynamiCrafter, and then continue to train it with another 40K iterations using \texttt{human} category data in the constructed dataset with \emph{dynamic wording} as text prompts. The baseline model is our DynamiCrafter, trained for 100K iterations. In addition, we maintain all other settings identical for fair comparison. As mentioned in Section 5 of the main paper, we use CLIP-SIM to evaluate the performance of DynamiCrafter$_\text{DCP}$, considering that CLIP is an open-domain text-image representation learner and is capable of associating the dynamics in the image with the appropriate dynamic wording.

\section{Other Controls}
\label{sec:control}

\subsection{FPS Control}
Since our model is also conditioned on FPS and trained with dynamic FPS, \ie 5--30, it is capable of generating image animations with varying motion magnitudes, as demonstrated in Figure~\ref{fig:fps_control}, where we show the results with `low FPS' and `high FPS' for simplicity.

\begin{figure}[!t]
    \centering
    \includegraphics[width=1\linewidth]{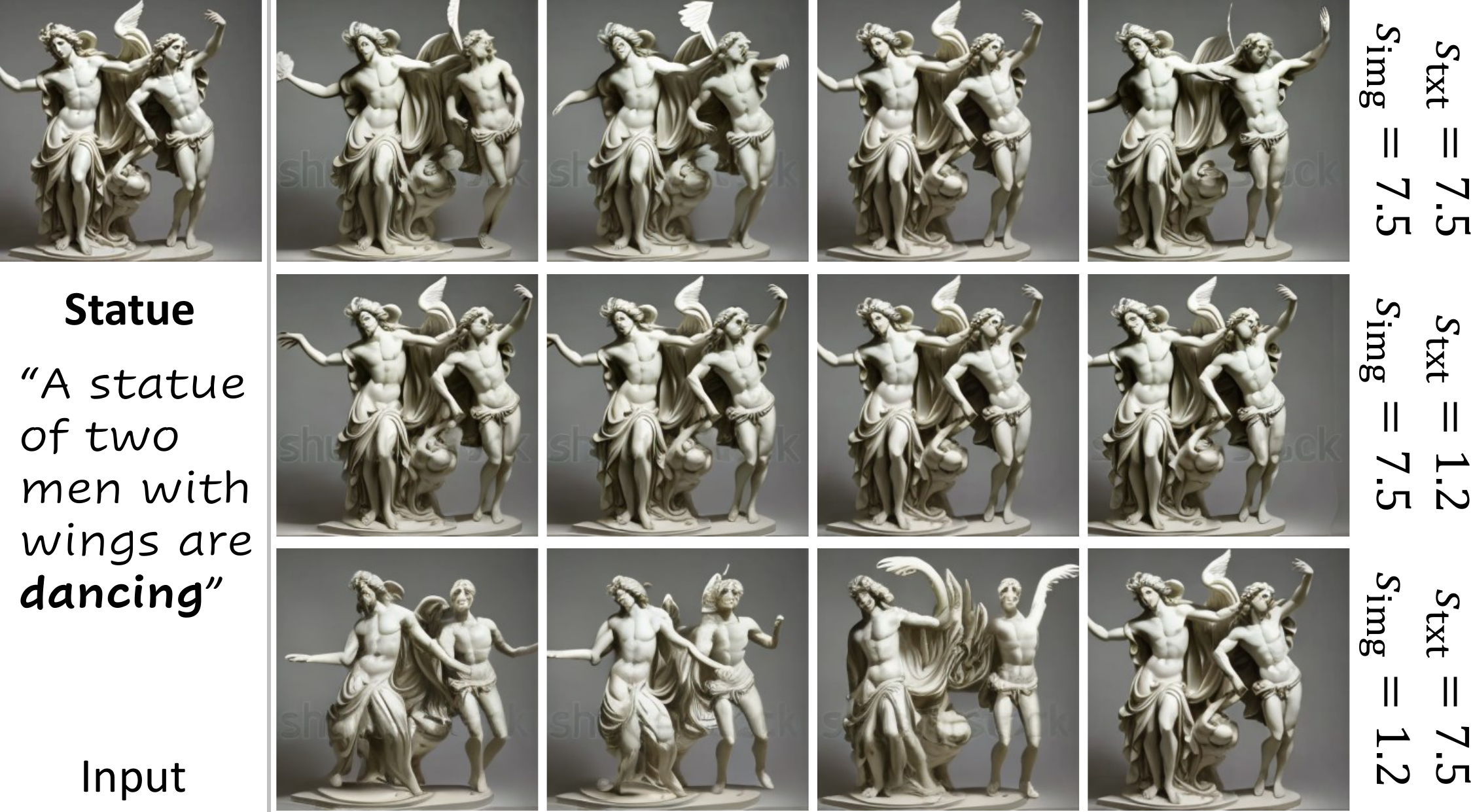}
    \caption{Visual comparisons of image animation results produced by various combinations of $s_\text{img}$ and $s_\text{txt}$.}
    \label{fig:cfg}
\end{figure}
\begin{figure}[!t]
    \centering
    \includegraphics[width=1\linewidth]{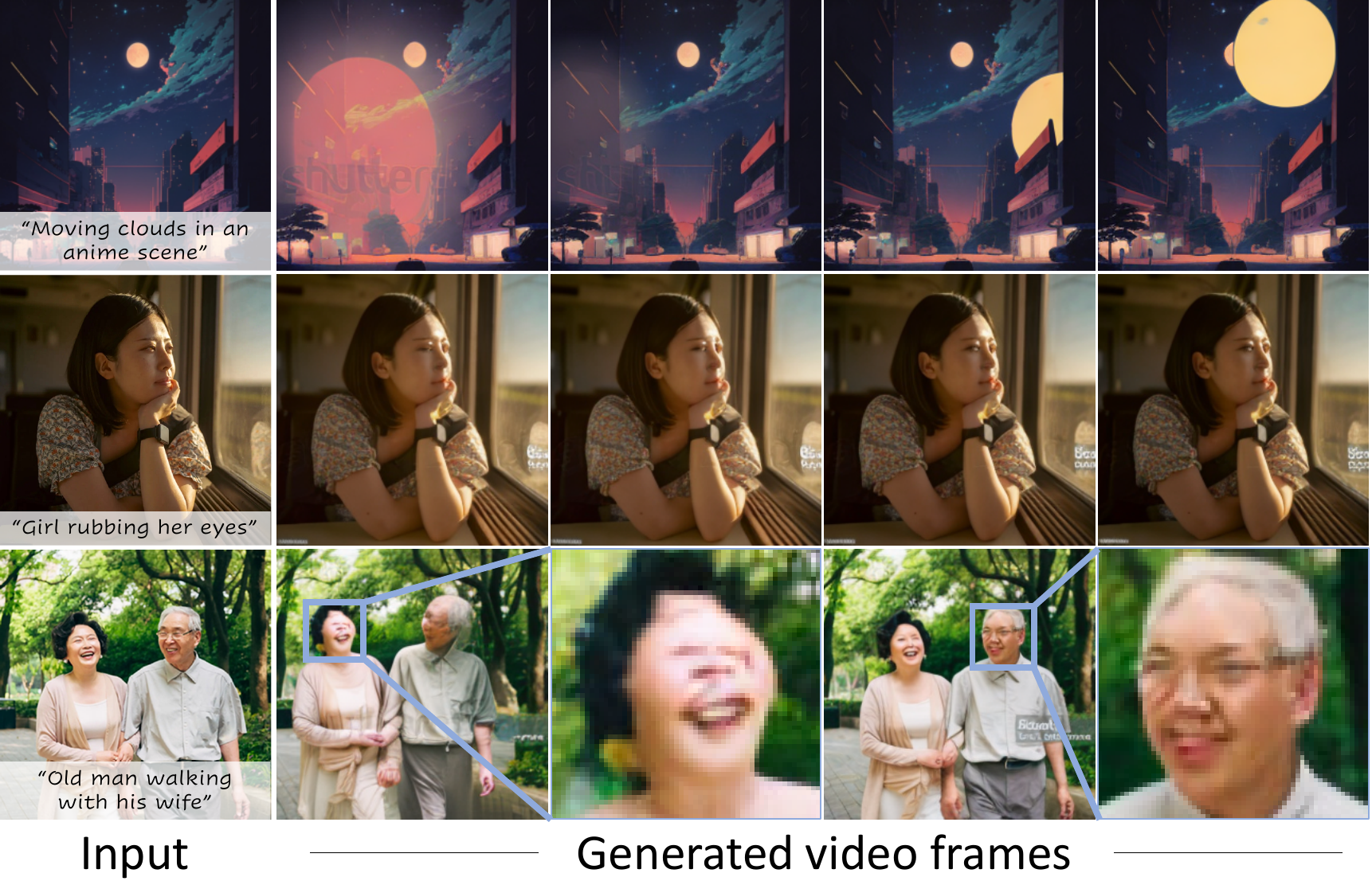}
    \caption{Failure cases of the challenging input condition in terms of semantic understanding (top), specific motion control with text (middle) and face distortion (bottom).}
    \label{fig:limitation}
\end{figure}

\begin{figure*}[!t]
    \centering
    \includegraphics[width=0.7\linewidth]{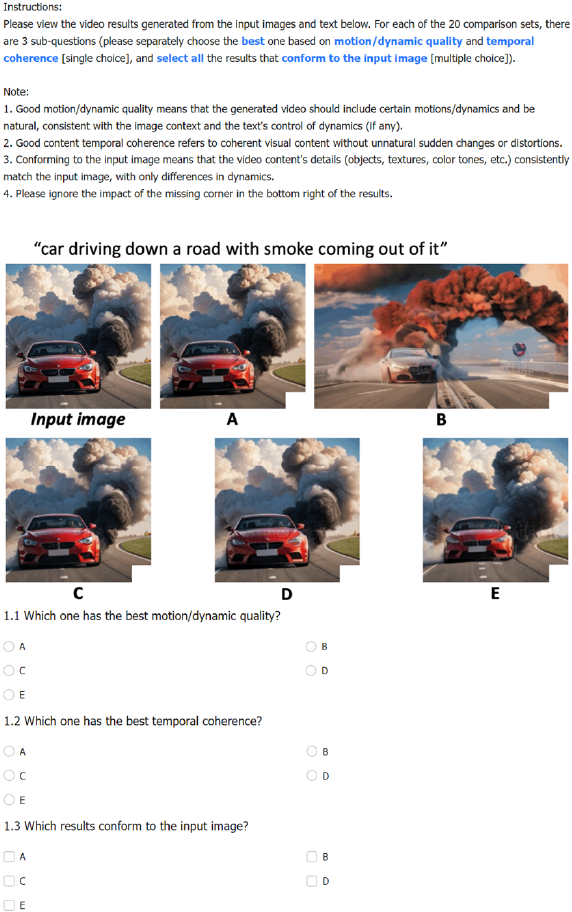}
    \caption{Designed user study interface. Each participant is required to evaluate 20 video comparisons and respond to three corresponding sub-questions for each comparison. Only one video is shown here due to the page limit.}
    \label{fig:user_study_screenshot}
\end{figure*}
\begin{figure*}[!t]
    \centering
    \includegraphics[width=\linewidth]{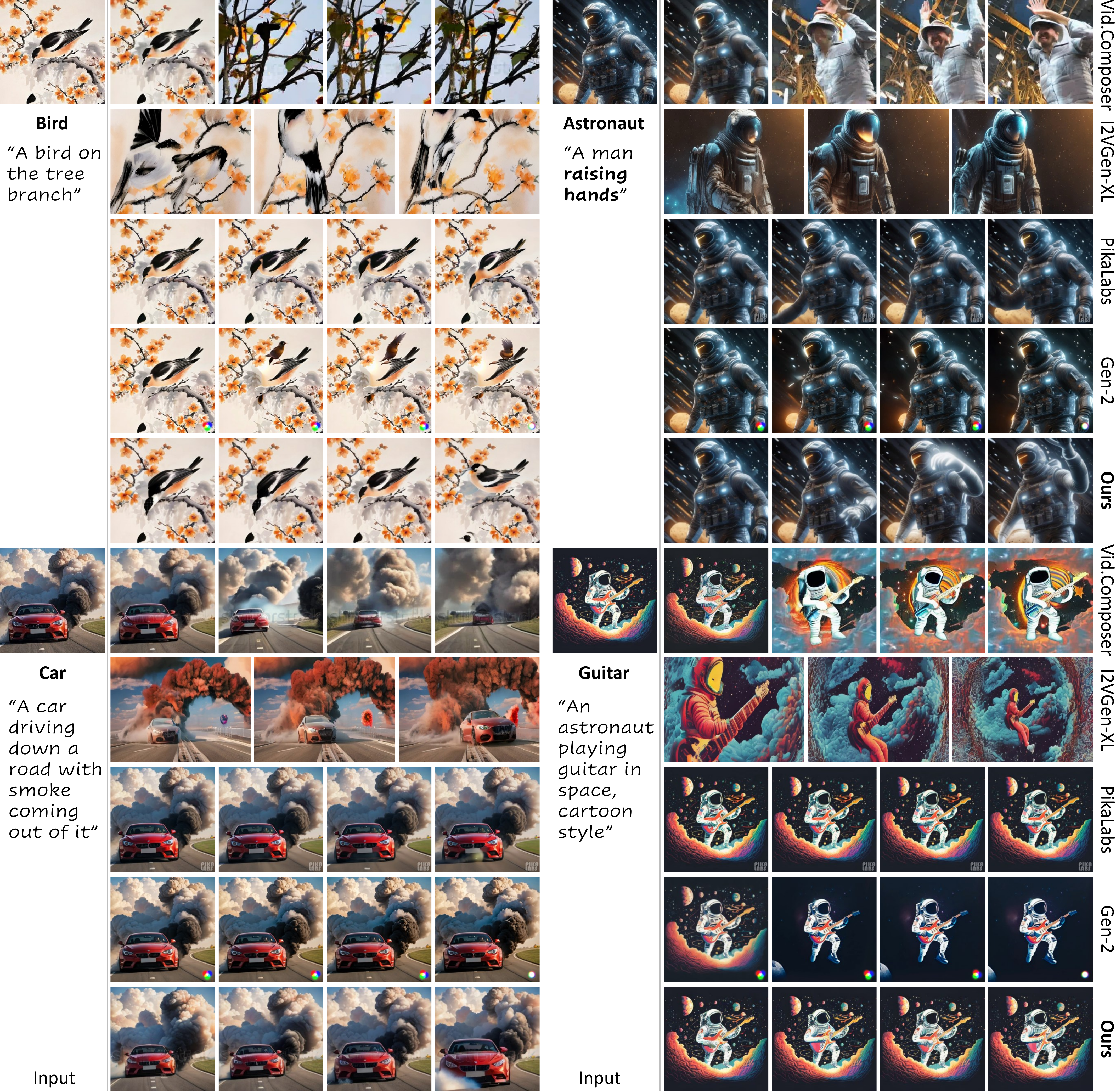}
    \caption{Visual comparisons of image animation results from VideoComposer, I2VGen-XL, PikaLabs, Gen-2, and our \emph{DynamiCrafter}.}
    \label{fig:more_1}
\end{figure*}

\begin{figure*}[!t]
    \centering
    \includegraphics[width=0.95\linewidth]{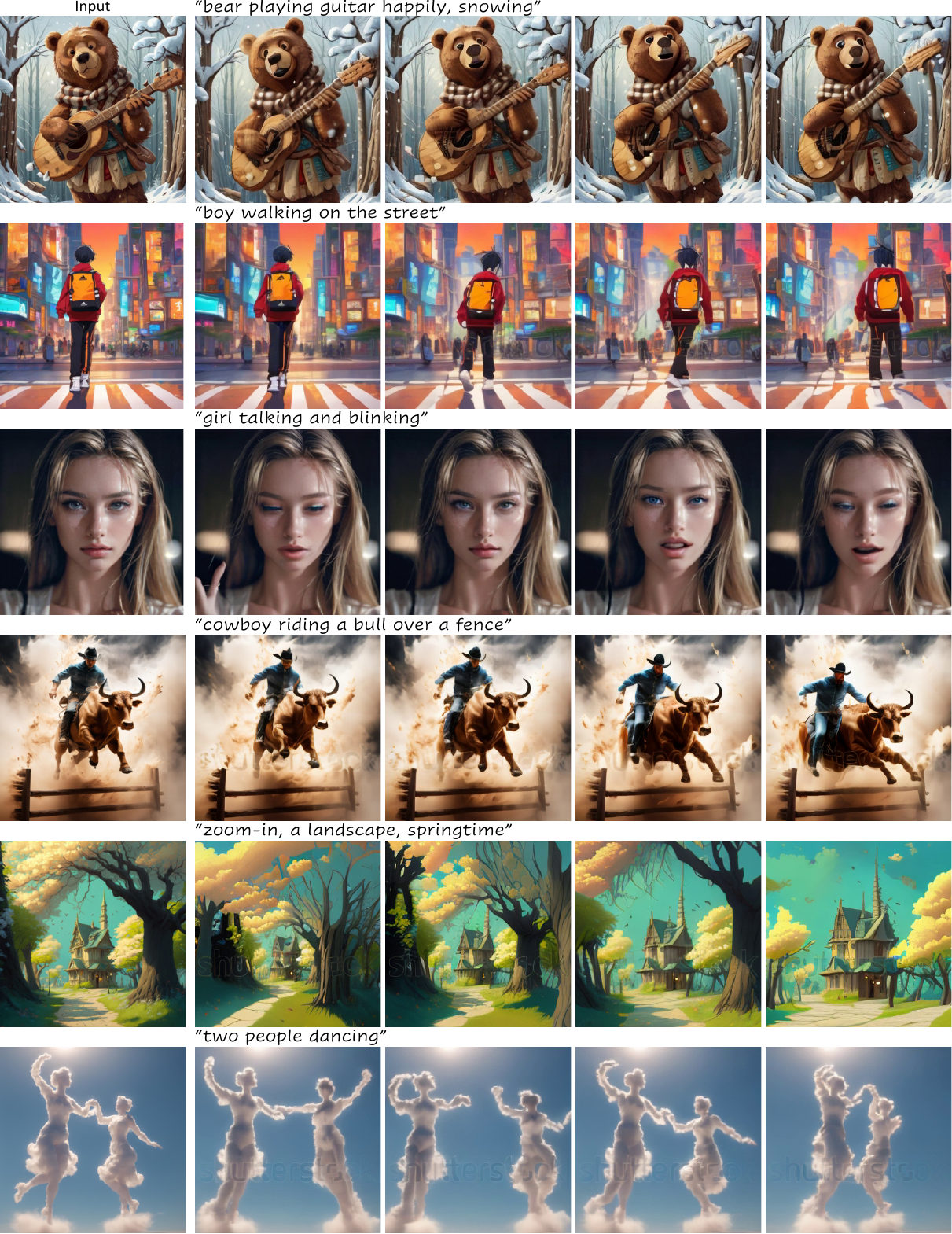}
    \caption{Gallery of our image animation results.}
    \label{fig:more_2}
\end{figure*}
\begin{figure*}[!t]
    \centering
    \includegraphics[width=0.95\linewidth]{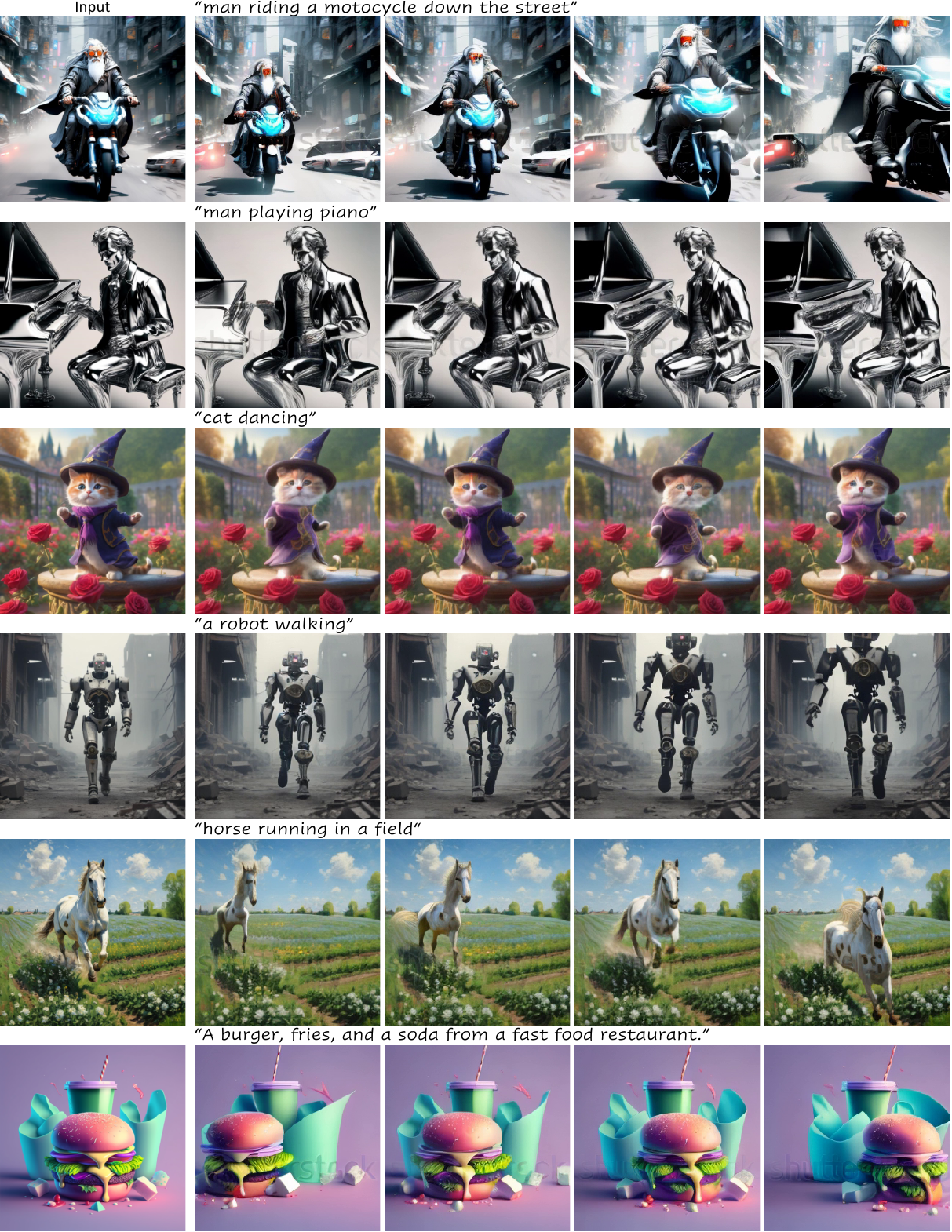}
    \caption{Gallery of our image animation results.}
    \label{fig:more_3}
\end{figure*}

\subsection{Multi-condition Classifier Free Guidance}
During inference, we adopt DDIM with multi-condition classifier guidance~\cite{ho2022classifier} and can adjust the introduced two guidance scales $s_\text{img}$ and $s_\text{txt}$ to trade off the impact of two control signals, as mentioned in Section 4.1 of the main paper. Specifically, it will affect how strongly the generated samples correspond with the input image and how strongly they correspond with the text prompt. Here we present the visual comparisons in Figure~\ref{fig:cfg}. In most cases, setting $s_\text{img}=s_\text{txt}=7.5$ works well, as the generated animations can well adhere to the input image and reflect the text prompt, as shown in Figure~\ref{fig:cfg}(top). By decreasing $s_\text{txt}$, the animation results tend to ignore the text condition, \eg, ``dancing", as shown in Figure~\ref{fig:cfg}(middle). Conversely, if $s_\text{img}$ is reduced, the results may not conform to the input image but well reflect the text prompt (see Figure~\ref{fig:cfg}(bottom)). This multi-condition classifier guidance offers greater flexibility based on user requirements.

\section{Limitations}
\label{sec:limitation}

Our approach is limited in several ways. Firstly, if the input image condition cannot be semantically understood, our model might struggle to produce convincing videos. Secondly, although we construct a dataset to improve motion control with text, which still lacks precise motion descriptions, rendering the inability to generate specific motions. Additionally, we adopt the LatentVDM pre-trained at low resolutions and with short durations due to limited computational resources, resulting in inheriting its slight flickering artifacts in high-frequency regions (see supplemental video results) and human face distortion issues, which are technically caused by the frame-wise VAE decoding. Thus the resultant frame quality of our method (such as resolution and fidelity) and video length may limit practical applications. Consequently, our method may not be ready for product (in contrast to commercial products like PikaLabs and Gen-2). Figure~\ref{fig:limitation} shows the examples of the mentioned failure cases. We leave these directions as future works.

\section{More Qualitative Results}
\label{sec:more}
\paragraph{More qualitative comparisons.}
In addition to Figure~4 in the main paper, we provide more qualitative comparisons in Figure~\ref{fig:more_1}. Consistent with the observations in the main paper, VideoComposer struggles to produce coherence video frames and tends to be misled by the text prompt. I2VGen-XL fails to preserve the local visual details of the input image and can only generate animations that semantically resemble the input. PikaLabs tends to generate still videos or videos with limited dynamics. Gen-2 may incorrectly interpret the given image, rendering unreasonable results and temporal inconsistency (as seen in the `Bird' and `Guitar' cases). Moreover, these baseline methods have difficulty considering the text prompt for motion control (\eg, \emph{raising hands} in the `Astronaut' case). In contrast, our approach can produce image animations with natural dynamics, better adherence to the input image, and motion control guided by the text prompt.

\paragraph{Gallery of our results.} We show more image animation results produced by our method in Figure~\ref{fig:more_2} and Figure~\ref{fig:more_3}. We collect those input images from the Internet, DAVIS~\cite{davis}, and JourneyDB~\cite{pan2023journeydb}.

\paragraph{Video results.}
We provide the video result at \url{https://doubiiu.github.io/projects/DynamiCrafter}. It contains the following parts: i). Showcases produced by our method, ii). Comparisons with baseline methods, iii). Motion control using text, iv). Applications, v). Other controls, vi). Ablation study, and vii). Limitations.

%% file: supp/hyper-parameters.tex
\begin{table}[!ht]
\setlength{\tabcolsep}{4pt}
\caption{Hyperparameters for our \emph{DynamiCrater}.}
\centering
\resizebox*{!}{.95\textheight}{
\setlength{\tabcolsep}{5pt}
    \begin{tabular}{lc}
    \toprule
    \textbf{Hyperparameter} & \textbf{\emph{DynamiCrafter}}   \\
    \midrule\midrule
    \textbf{Spatial Layers}  &  \\
    \textit{Architecture} \\
    LDM & \cmark  \\
    $f$ & 8 \\
    $z$-shape & $32 \times 32 \times 4$  \\
    Channels & 320 \\
    Depth & 2 \\
    Channel multiplier & 1,2,4,4\\
    Attention resolutions & 64,32,16\\
    Head channels & 64 \\
    Input channels & 8 \\
    Output channels & 4 \\
    \midrule
    \emph{Dual-CA Conditioning} & \\
    Embedding dimension & 1024 \\
    CA resolutions & 64,32,16\\
    CA txt sequence length & 77\\
    CA ctx sequence length & 16\\
    \midrule
    \emph{FPS Conditioning} & \\
    Embedding dimension & 1280 \\
    FPS sampling range & 5--30\\
    \midrule
    \emph{Concat Conditioning} & \\
    Embedding dimension & 4 \\
    Index of video frame & Random \\
    Extension in temporal dim. & Repeat \\
    \midrule\midrule
    \textbf{Temporal Layers}  & \\
    \textit{Architecture}  &  \\
    Transformer depth & 1 \\
    Attention resolutions & 64,32,16 \\
    Head channels & 64\\
    Positional encoding &\xmark\\
    Temporal conv layer num & 4 \\
    Temporal kernel size & 3,1,1 \\
    \midrule \midrule
    \textbf{Training}  &  \\
    Parameterization & $\boldsymbol{\varepsilon}$ \\
    Learnable para. & Spatial layers\\
                  & $\mathcal{P}$ with ctx CA \\
    \# train steps  & 100K   \\
    Learning rate & $5\times 10^{-5}$  \\
    Batch size per GPU  & 8\\
    \# GPUs  & 8 \\
    GPU-type  & V100-32GB \\
    Sequence length & 16\\
    \midrule\midrule
    \textbf{Diffusion Setup}  &   \\
    Diffusion steps & 1000 \\
    Noise schedule & Linear  \\
    $\beta_{0}$ & 0.00085\\
    $\beta_{T}$ & 0.0120\\
    \midrule\midrule
    \textbf{Sampling Parameters}  &  \\
    Sampler  &  DDIM   \\
    Steps  & 50\\
    $\eta$  & 1.0 \\
    Guidance scale $s_\text{txt}$ & 7.5 \\
    Guidance scale $s_\text{img}$ & 7.5 \\
    \bottomrule
\end{tabular}%
}
\label{table:hyperparameters}
\vspace{-0.4cm}
\end{table}